\pdfoutput=1

\documentclass[11pt]{article}

\usepackage{acl}
\usepackage{times}
\usepackage{latexsym}
\usepackage[T1]{fontenc}
\usepackage[utf8]{inputenc}
\usepackage{microtype}
\usepackage{inconsolata}
\usepackage{graphicx}

\usepackage{enumitem}
\usepackage{soul}
\usepackage{amsmath}
\usepackage{amsfonts}
\usepackage{bbm}
\usepackage{booktabs}
\usepackage{multirow}
\usepackage{paralist}
\usepackage{xspace}
\usepackage{subcaption}
\usepackage{colortbl}
\usepackage{tabularx}
\usepackage{makecell}
\usepackage{listings}

\newcommand{\bscom}[3][]{
    \noindent
    \st{#2}
    {\selectfont\color{blue}#3}
    \ifx\\#1\\
    \else
        {\selectfont\color{violet}[#1]}
    \fi
}

\newcommand{\rouge}{\textsc{ROUGE}\xspace}
\newcommand{\rougen}[1]{\textsc{ROUGE}$_{\text{#1}}$\xspace}
\newcommand{\summac}{\textsc{SummaC}\xspace}
\newcommand{\alignscore}{\textsc{AlignScore}\xspace}

\newcommand{\bert}{\textsc{BERTScore}\xspace}

\newcommand{\bleurt}{\textsc{BLEURT}\xspace}
\newcommand{\llmcov}{LLM-Coverage\xspace}

\newcommand{\llmfaith}{LLM-Faithfulness\xspace}

\newcommand{\polisum}{\textsc{PoliSum}\xspace}

\newcommand{\llamathree}{\texttt{Llama-3.1-8B-Instruct}\xspace}
\newcommand{\mistral}{\texttt{Mistral-7B-Instruct-v0.3}\xspace}
\newcommand{\qwenfourteen}{\texttt{Qwen2.5-14B-Instruct}\xspace}

\newcommand{\fixnum}[1]{\makebox[1.2cm][l]{#1}}
\newcolumntype{P}[1]{>{\raggedright\arraybackslash\vspace{-1em}}p{#1}}
\newcommand{\hlcolor}[3]{{\sethlcolor{#2!#3}\hl{#1}}}

\title{Reranking-based Generation for Unbiased Perspective Summarization}

\author{Narutatsu Ri \and 
        Nicholas Deas \and 
        Kathleen McKeown \\ 
        Department of Computer Science, Columbia University \\
	{\tt \{wl2787, nid2107, km\}@columbia.edu}
}

\begin{document}
\maketitle

\begin{abstract}
\vspace{-3mm}
Generating unbiased summaries in real-world settings such as political perspective summarization remains a crucial application of Large Language Models (LLMs).
Yet, existing evaluation frameworks rely on traditional metrics for measuring key attributes such as coverage and faithfulness without verifying their applicability, and efforts to develop improved summarizers are still nascent.
We address these gaps by 
\begin{inparaenum}[(1)]
    \item  identifying reliable metrics for measuring perspective summary quality, and
    \item investigating the efficacy of LLM-based methods beyond zero-shot inference.
\end{inparaenum}
Namely, we build a test set for benchmarking metric reliability using human annotations and show that traditional metrics underperform compared to language model–based metrics, which prove to be strong evaluators. 
Using these metrics, we show that reranking-based methods yield strong results, and preference tuning with synthetically generated and reranking-labeled data further boosts performance.
Our findings aim to contribute to the reliable evaluation and development of perspective summarization methods.
\end{abstract}
\vspace{-6mm}
\section{Introduction} 
\label{sec:introduction}
\vspace{-2mm}

Article summarization is a key application of Large Language Models (LLMs) given their recent breakthroughs in text generation capabilities \cite{goyal2023news, zhang2024benchmarking}.
Critically, however, LLMs often exhibit undesirable behaviors and input-level biases toward spurious features (e.g., position) \citep{jung2019earlier, chhabra2024revisiting, liu2024lost}, resulting in unbalanced input coverage \citep{zhang2024fair} and hallucination \citep{maynez2020faithfulness}.
These issues are especially problematic in opinionated article summarization \citep{amplayo2021aspect, iso2022comparative}, where unbiased representation of diverse viewpoints is crucial.

Recent studies in opinion summarization address these risks by developing tasks and methods that generate summaries free of framing bias \cite{lee2022neus}, fairly represent input diversity \cite{zhang2024fair, feng2024modular}, or preserve the source perspectives \cite{lei2024polarity, liu2024p3sum}. 
Within this domain, \emph{perspective summarization} \cite{deas2025summarization} serves as a representative evaluation setting, where models are tasked to generate precise, perspective-specific summaries from multi-document inputs containing diverse political views.
However, two gaps remain unaddressed in this setting: 
\begin{inparaenum}[(1)]
    \item existing evaluation metrics are primarily derived from news summarization domains and have not been validated for measuring \emph{perspective} summary quality, and 
    \item the effectiveness of LLM-based methods beyond zero-shot inference in generating unbiased, high-quality perspective summaries remains underexplored.
\end{inparaenum}

To address these gaps, we first identify effective metrics for measuring summary quality by constructing a test set to evaluate existing metrics.
We focus on two key attributes that a desirable summary should have:
\emph{perspective coverage}---the extent to which the summary includes all key content from the intended perspective, and 
\emph{perspective faithfulness}---the degree to which the summary excludes content unsupported by the source articles of the target perspective. 
We collect key point annotations from articles to create controlled summaries with varied key point selections and assigned ground truth scores.
We find that language model-based metrics such as \alignscore \cite{zha2023alignscore} and prompting-based scoring \cite{zheng2023judging} serve as strong evaluators, whereas traditional metrics (\rouge \cite{lin2004rouge}, \bert \cite{zhang2020bertscore}) underperform.

Following this, we evaluate methods for generating perspective summaries with improved coverage and faithfulness beyond zero-shot inference. 
We benchmark prompting frameworks, mechanistic methods for mitigating input biases, and reranking-based methods that select the best candidate based on proxy metrics.
Using both human and automatic evaluations, we show that reranking outperforms zero-shot inference and prompting-based methods, while prompting only yields marginal improvements over zero-shot inference.
Notably, preference tuning with Direct Preference Optimization (DPO) \cite{rafailov2023direct} on reranked generations further boosts performance on both attributes and particularly improving faithfulness.
Our results suggest that current LLMs can generate high-quality perspective summaries with strong coverage and faithfulness, and that preference-based training can further boost performance.

In summary, our contributions are as follows:
\begin{compactitem} 
    \item We construct a controlled test set and identify effective metrics for measuring coverage and faithfulness for perspective summarization. 
    \item We evaluate various generation methods and demonstrate that reranking-based approaches deliver the best performance in producing summaries with improved coverage perspective and faithfulness.
    Notably, preference tuning on reranked generations significantly improves both attributes, with the most pronounced gains in faithfulness.
    \item We conduct ablation studies and show that commonly employed prompting frameworks consistently underperform relative to reranking-based methods, even when scaled to high-resource settings.\footnote{Our code is available at \url{https://github.com/narutatsuri/Unbiased-Perspective-Summarization}.}
\end{compactitem}

\vspace{-3mm}
\section{Related Work}
\label{sec:related_work}
\vspace{-2mm}

\paragraph{Summary Evaluation.}
Summary evaluation traditionally relies on reference-based metrics, including $n$-gram-based methods (\rouge \cite{lin2004rouge}, BLEU \cite{papineni-bleu}, CHRF \cite{popovic-2015-chrf}), model-based coverage scores (\bert \cite{zhang2020bertscore}, \bleurt \cite{sellam2020bleurt}), and composite measures (METEOR \cite{banerjee-meteor}). 
In response to unreliable references, recent work proposes reference-free metrics that target aspects such as faithfulness and factual consistency. 
Neural approaches dominate this space, including end-to-end classifiers (FactCC \cite{kryscinski-factcc}), QA-based methods (QAGS \cite{wang-qags}, QAFactEval \cite{fabbri-qafacteval}), NLI models (\summac \cite{laban2022summac}), and information alignment models (\alignscore \cite{zha2023alignscore}). 
Here, we focus on automatic, reference-free measures of coverage and faithfulness, but conduct a novel evaluation of their reliability in a multi-document perspective summarization task.

Beyond developing improved faithfulness metrics, prior works focus on improving the factual consistency of summarizers, with studies noting the tradeoff between abstractiveness and faithfulness \cite{durmus-feqa, dreyer-tradeoff}. 
Accordingly, some methods improve faithfulness without increasing extraction \cite{ladhak-faithful}, while others modify training via contrastive \cite{nan-contrastive}, multi-task \cite{chen-multi}, or reinforcement learning \cite{roit-rlef} methods. 
In contrast, we show that reranking-based methods serve as a strong baseline that yields high faithfulness without sacrificing abstractiveness, and a DPO-based approach trained on reranked self-generated summaries further improves both qualities.

\vspace{-2mm}
\paragraph{Perspective-Conditioned Summarization.}
Existing research on opinion summarization and related tasks has primarily focused on domains such as product reviews \cite{brazinskas-opinion}, while recent work has broadened to a range of tasks on opinionated texts. 
Most single-document methods aim to preserve authorial intent \cite{liu2024p3sum} or polarity \cite{lei2024polarity}, whereas multi-document summarization must integrate varied perspectives.
For instance, \citet{lee-neus} generates politically neutral summaries from sets of left-, right-, and center-leaning news articles.
Other approaches aim to fairly represent diverse perspectives in reviews \cite{zhang2024fair}, controllably represent community perspectives \cite{feng2024modular}, generate consensus summaries \cite{bakker2022finetuning}, or produce multiple summaries reflecting distinct political perspectives \cite{deas2025summarization}. 
In line with these works, we summarize the political perspective among a set of input passages while addressing the coverage and faithfulness issues observed in existing models as highlighted in these studies.
\vspace{-3mm}
\section{Measuring Summary Quality}
\label{sec:metric_definition_and_evaluation}
\vspace{-3mm}

In perspective summarization, the summarizer is given two perspectives $\theta_1, \theta_2$, each with a source article $D_{t, \theta}, \theta \in \{\theta_1, \theta_2\}$, comprising a set of documents $D_{t, \theta} = \{d_{t, \theta}^{(i)} \mid i \in \mathbb{N}\}$, that present opinions on topic $t$.
We study the setting where the summarizer is tasked to generate a summary that encapsulates all \emph{key points} directly supporting a specified perspective's stance.
Concretely, a high-quality perspective summary should:
\begin{inparaenum}[(1)]
    \item include all key points from each relevant document, and
    \item avoid including any content unsupported by or in opposition to the perspective's documents.
\end{inparaenum}
We formalize these properties as follows: 
\begin{compactitem}
    \item \emph{Perspective Coverage}: The ratio of key points included in the summary relative to the total number of key points.
    \item \emph{Perspective Faithfulness}: The ratio of relevant key points included in the summary relative to the total number of included key points.
\end{compactitem}
Although metrics for similar properties exist in other summarization domains, it is unclear whether they effectively measure the properties as defined above for the perspective summarization task.
We therefore evaluate how well these metrics capture our definitions of coverage and faithfulness.\footnote{We note that our notions of coverage and faithfulness differ from prior work \citep{zhang2021finding, tang2024minicheck, song2024finesure}, as we assess both attributes with respect to the correct inclusion of key points.
For brevity, we use perspective coverage and faithfulness interchangeably with coverage and faithfulness.}

\vspace{-2mm}
\subsection{Assessing Metric Quality}
\label{sec:assessing_metric_quality}
\vspace{-1mm}

Quantifying the efficacy of existing metrics requires article-summary pairs with ground truth scores for evaluation.
Although perspective summarization datasets such as \polisum \citep{deas2025summarization} include reference summaries, each document is paired with only one summary without assigned scores for coverage and faithfulness.
Hence, we construct a test set of article-summary pairs with assigned ground truth scores for coverage and faithfulness and evaluate how well existing metrics align with these scores.

\begin{figure}[t]
    \centering
    \includegraphics[width=\columnwidth]{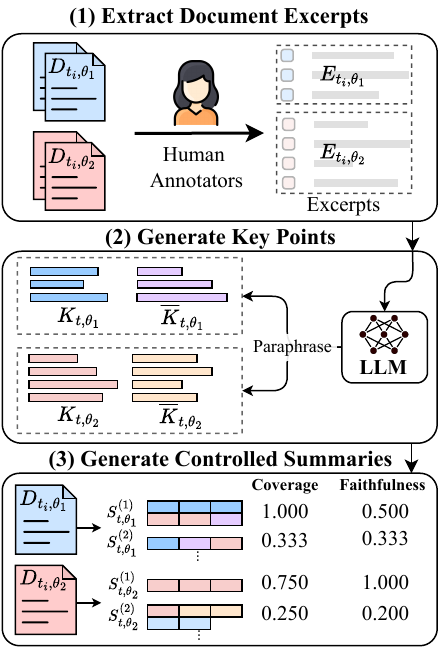}
    \vspace{-8mm}
    \caption{Pipeline for curating the synthetic testbed for metric evaluation. 
    Annotators extract the most important excerpts $E_{t,\theta}$ from articles $D_{t,\theta}$, which are paraphrased into key points $K_{t,\theta}$ and adversarial key points $\overline{K}_{t,\theta}$. 
    We then curate summaries with a diverse range of coverage and faithfulness scores using the key points.}
    \vspace{-4mm}
    \label{fig:testbed_curation}
\end{figure}

\begin{table}[t]
    \centering
    \scriptsize
    \begin{tabular}{p{0.2\columnwidth}|p{0.67\columnwidth}}
        \toprule
        \bf Article Topic & Ron DeSantis \\
        \midrule
        \bf \makecell[tl]{\vspace*{0pt}Perspective\\Source Article\\(Key Points)} & 
        \hlcolor{DeSantis has shown authoritarian tendencies throughout his time in office.}{green}{20}\newline
        \hlcolor{DeSantis' election police proposal chills legitimate election work and threatens democracy.}{blue}{20}\newline
        \hlcolor{DeSantis' claim that Florida is the freest state contradicts restrictions on health, protest, and education.}{violet}{30} \\
        \midrule      
        \midrule      
        \bf \makecell[tl]{\vspace*{0pt}Synthetic\\(High-Quality)} & The article contends that \hlcolor{DeSantis's proposal for an election police squad undermines legitimate election activities and democracy}{blue}{20}, \hlcolor{contradicts his claim of Florida being the freest state by restricting various freedoms}{violet}{30}, and \hlcolor{highlights his persistent authoritarian inclinations during his tenure}{green}{20}. \\
        \midrule      
        \bf \makecell[tl]{\vspace*{0pt}Synthetic\\(Low-Quality)} & The article highlights \hlcolor{DeSantis's authoritarian tendencies and his contradiction in calling Florida the freest state while restricting freedoms}{violet}{30}, but \hlcolor{praises his election police proposal for protecting elections and strengthening democracy}{red}{30} and \hlcolor{urges Trump to prioritize GOP leadership in Florida and retaking the House over personal pride}{orange}{30}. \\
        \bottomrule
    \end{tabular}
    \vspace{-3mm}
    \caption{Examples of constructed summaries. 
    For brevity, only curated key points are shown for the source article.
    \hlcolor{Purple}{violet}{20}, \hlcolor{blue}{blue}{20}, and \hlcolor{green}{green}{20} highlights denote relevant key points, while \hlcolor{red}{red}{30} and \hlcolor{orange}{orange}{30} highlights respectively indicate adversarial and opposite key points.}
    \label{tab:metric_example_summaries}
    \vspace{-5mm}
\end{table}

\vspace{-2mm}
\paragraph{Test Set Construction.}
To assign meaningful ground truth scores for both attributes, we identify all key points in an article and create summaries using different combinations of these points. 
We begin with articles from \polisum\footnote{In the \polisum dataset, $|D_{t, \theta}|$ has a mean of $5.31$ with a standard deviation of $1.45$.} and collect human annotations in which annotators highlight document excerpts supporting the perspective's stance. 
See \S\ref{app:annotation_interfaces} for the annotation interface.

Formally, given an article $D_{t, \theta}$, we collect a set of excerpts $E_{t, \theta}$ defined as:
\vspace{-2mm}
\begin{equation*}
    \resizebox{\hsize}{!}{$
    E_{t, \theta} = \{e_{t, \theta}^{(i)} \mid e_{t, \theta}^{(i)} \subseteq d_{t, \theta}^{(i)}, d_{t, \theta}^{(i)} \,\,\textup{contains a key point}\},
    $}
\end{equation*}
where $|E_{t, \theta}| \leq |D_{t, \theta}|$ (i.e., not all documents contain key points, and each document has at most one key point).
As an excerpt $e_{t, \theta}^{(i)}$ may not clearly convey the main argument, we use an LLM $f: E_{t,\theta}\rightarrow K_{t,\theta}$ to rewrite excerpts into key points to form the set $K_{t, \theta}$:\footnote{See \S\ref{app:paraphrasing} for further details.}
\vspace{-2mm}
\begin{equation*}
    K_{t, \theta} = \{k_{t,\theta}^{(i)} \mid k_{t,\theta}^{(i)} = f(e_{t,\theta}^{(i)}),\,\,e_{t,\theta}^{(i)} \in E_{t,\theta}\}.
    \vspace{-2mm}
\end{equation*}
Given $K_{t, \theta}$, we construct summaries $S_{t,\theta}^{(i)}$ by selecting $k_g$ key points from $K_{t, \theta}$ and $k_b$ from a set of unfaithful key points.
We generate unfaithful key points by sampling key points from the opposing perspective (e.g., using key points from the left-leaning document for right-perspective summaries), and by reversing the content of key points in $K_{t, \theta}$ to form adversarial key points $\overline{K}_{t, \theta}$ \citep{laban2022summac}.
We then define:
\vspace{-2mm}
\begin{align}
    \textup{Coverage}(S_{t,\theta}^{(i)}) &= \frac{k_g}{|K_{t, \theta}|}, \label{eqn:coverage}\\
    \textup{Faithfulness}(S_{t,\theta}^{(i)}) &= \frac{k_g}{k_g + k_b}. \label{eqn:faithfulness}
\end{align}
We provide examples of summaries with varying scores in Table~\ref{tab:metric_example_summaries}.
With this procedure, we produce summaries with error levels ranging from few minor omissions to many faithfulness errors. 
We collect annotations for $50$ documents from $5$ annotators and generate a varying number of summaries for each document, ultimately curating $370$ article-summary pairs in total.
We illustrate the process in Figure~\ref{fig:testbed_curation}.
See \S\ref{app:annotation_details} for further annotation details.

\begin{figure}[t]
    \centering
    \begin{minipage}{\columnwidth}
        \begin{lstlisting}[basicstyle=\ttfamily\scriptsize, 
                           frame=single, 
                           breaklines=true,
                           breakatwhitespace=false,
                           columns=fullflexible,
                           keepspaces=true,
                           xleftmargin=0pt,
                           xrightmargin=0pt,
                           breakindent=0ex]
You are an evaluator. Your task is to determine how well a generated summary captures all of the main arguments from a source article. This is a measure of "coverage," which does not necessarily address factual accuracy (faithfulness) but focuses on completeness of content. The scale for coverage is:
1. No Coverage: The summary does not include any of the main arguments from the article.
2. Low Coverage: The summary includes only a few of the main arguments from the article, omitting most.
3. Medium Coverage: The summary contains around half of the article's main arguments.
4. High Coverage: The summary contains most of the main arguments from the article, missing only a few.
5. Perfect Coverage: The summary includes all major points mentioned in the article, leaving out nothing important.

Follow these steps carefully:
(Omitted for Brevity)

# Source Article:
(article)
# Summary:
(summary)
# Coverage Score (1~5 only):
        \end{lstlisting}
    \end{minipage}
    \vspace{-5mm}
    \caption{Example prompt for \llmcov. We follow the prompt instruction format in \citet{wu2024less}. 
    Portions of the prompt are omitted for brevity.
    See \S\ref{app:metric_annotation_prompts} for complete prompt instructions.}
    \vspace{-5mm}
    \label{fig:metric_example_prompt}
\end{figure}

\vspace{-1mm}
\paragraph{Benchmarked Metrics.}
As baselines, we respectively use the recall and precision variants of \textbf{\rouge} \cite{lin2004rouge} and \textbf{\bert} \citep{zhang2020bertscore} for measuring coverage and faithfulness.
We also report \textbf{BLEURT} \cite{sellam2020bleurt} as an additional coverage metric.
For faithfulness, we test \textbf{\summac} \cite{laban2022summac} (NLI-based inconsistency detection metric), \textbf{\alignscore} \cite{zha2023alignscore} (factual consistency metric), the consistency dimension of \textbf{UniEval} \cite{zhong2022towards} (T5-based multi-task evaluator), \textbf{MiniCheck} \cite{tang2024minicheck} (\textsc{Flan-T5} model for fact-checking via entailment), and the faithfulness dimension of \textbf{FineSurE} \cite{song2024finesure} (span-level fact verification).
See \S\ref{app:metric_configurations} for details on metric configurations and model checkpoints.

Furthermore, recent studies suggest that LLMs serve as effective evaluators \cite{chiang2023large, dubois2023alpacafarm, chen2023exploring}, including for some dimensions of summary qualities \cite{jain2023multi, wu2024less}.
Hence, we examine two LLM-as-a-Judge settings where the source article and generated summary are passed as input alongside tailored prompts \citep{liu2023g}.
We respectively term these \textbf{\llmcov} and \textbf{\llmfaith} for convenience.
As an example, see Figure~\ref{fig:metric_example_prompt} for the \llmcov prompt instruction.
We use \mistral as the default backbone based on evaluation performance.
See \S\ref{app:llm_metric_benchmarking} for results using alternative models.

Note that, by our formulation, coverage corresponds to recall and faithfulness to precision in key point inclusion. 
Hence, we report results on both attributes for all metrics and show that recall-based metrics do not capture faithfulness and vice versa to verify the reliability of our curated test set.

\begin{table}[t]
    \centering
    \resizebox{\columnwidth}{!}{
    \begin{tabular}{l|cc|cc}
    \toprule
     & \multicolumn{2}{c|}{\bf Coverage} & \multicolumn{2}{c}{\bf Faithfulness} \\
    \bf Metric & \bf Corr.~($\rho_s$) & \bf Winrate & \bf Corr.~($\rho_s$) & \bf Winrate \\
    \midrule\midrule
    \rougen{L} ($R$) 
      & \cellcolor{orange!47} \fixnum{$0.473^{***}$} 
      & \cellcolor{orange!60} $0.780 \pm \text{\small 0.048}$ 
      & \cellcolor{orange!0}  \color{gray}\fixnum{$-0.038$} 
      & \cellcolor{orange!0}  \color{gray}$0.393 \pm \text{\small 0.063}$ \\
    \bert ($R$) 
      & \cellcolor{orange!45} \fixnum{$0.527^{***}$} 
      & \cellcolor{orange!60} $0.815 \pm \text{\small 0.018}$ 
      & \cellcolor{orange!10} \color{gray}\fixnum{$-0.032^{**}$} 
      & \cellcolor{orange!25} \color{gray}$0.415 \pm \text{\small 0.015}$ \\
    \bleurt 
      & \cellcolor{orange!8}  \fixnum{$0.086$} 
      & \cellcolor{orange!5}  $0.530 \pm \text{\small 0.067}$ 
      & \cellcolor{orange!0}  \color{gray}\fixnum{$-0.014$} 
      & \cellcolor{orange!5}  \color{gray}$0.527 \pm \text{\small 0.063}$ \\
    \llmcov 
      & \cellcolor{orange!70} \fixnum{$0.707^{***}$} 
      & \cellcolor{orange!45} $0.739 \pm \text{\small 0.047}$ 
      & \cellcolor{orange!39} \color{gray}\fixnum{$0.393^{***}$} 
      & \cellcolor{orange!10} \color{gray}$0.431 \pm \text{\small 0.115}$ \\
    \midrule
    \rougen{L} ($P$) 
      & \cellcolor{orange!0}  \color{gray}\fixnum{$-0.169^{**}$} 
      & \cellcolor{orange!0}  \color{gray}$0.443 \pm \text{\small 0.056}$ 
      & \cellcolor{orange!33} \fixnum{$0.333^{***}$} 
      & \cellcolor{orange!34} $0.714 \pm \text{\small 0.076}$ \\
    \bert ($P$) 
      & \cellcolor{orange!10} \color{gray}\fixnum{$0.073^{**}$} 
      & \cellcolor{orange!15} \color{gray}$0.510 \pm \text{\small 0.030}$ 
      & \cellcolor{orange!36} \fixnum{$0.366^{***}$} 
      & \cellcolor{orange!35} $0.655 \pm \text{\small 0.020}$ \\  
    \summac 
      & \cellcolor{orange!3}  \color{gray}\fixnum{$0.028$} 
      & \cellcolor{orange!0}  \color{gray}$0.491 \pm \text{\small 0.084}$ 
      & \cellcolor{orange!0}  \fixnum{$-0.016$} 
      & \cellcolor{orange!0}  $0.315 \pm \text{\small 0.066}$ \\
    \alignscore 
      & \cellcolor{orange!26} \color{gray}\fixnum{$0.261^{***}$} 
      & \cellcolor{orange!20} \color{gray}$0.503 \pm \text{\small 0.074}$ 
      & \cellcolor{orange!60} \fixnum{$0.650^{***}$} 
      & \cellcolor{orange!65} $0.773 \pm \text{\small 0.061}$ \\
    UniEval ($C$) 
      & \cellcolor{orange!27} \color{gray}$0.267^{***}$ 
      & \cellcolor{orange!55} \color{gray}$0.545 \pm \text{\small 0.055}$ 
      & \cellcolor{orange!63} $0.629^{***}$ 
      & \cellcolor{orange!77} $0.768 \pm \text{\small 0.054}$ \\
    MiniCheck 
      & \cellcolor{orange!10} \color{gray}$0.099$ 
      & \cellcolor{orange!44} \color{gray}$0.435 \pm \text{\small 0.066}$ 
      & \cellcolor{orange!58} $0.578^{***}$ 
      & \cellcolor{orange!75} $0.747 \pm \text{\small 0.074}$ \\
    FineSurE ($F$) 
      & \cellcolor{orange!27} \color{gray}$0.271^{***}$ 
      & \cellcolor{orange!29} \color{gray}$0.288 \pm \text{\small 0.076}$ 
      & \cellcolor{orange!8}  $0.084$ 
      & \cellcolor{orange!22} $0.216 \pm \text{\small 0.072}$ \\    
    \llmfaith 
      & \cellcolor{orange!46} \color{gray}\fixnum{$0.462^{***}$} 
      & \cellcolor{orange!40} \color{gray}$0.398 \pm \text{\small 0.055}$ 
      & \cellcolor{orange!71} \fixnum{$0.706^{***}$} 
      & \cellcolor{orange!54} $0.537 \pm \text{\small 0.091}$ \\
    \bottomrule
    \end{tabular}
    }
    \vspace{-3mm}
    \caption{Comparison of Spearman correlation ($\rho_s$) and Winrate with 95\% Confidence Interval (CI)  across all metrics. 
    Darker shading indicates better performance.
    Asterisks indicate significance levels ($^{*}, ^{**}, ^{***}$ for $p<0.05, 0.01, 0.001$, respectively).
    $P$ and $R$ denote the precision and recall variants of each metric. 
    Note the random baseline for Winrate is $0.500$.}
    \vspace{-5mm}
    \label{tab:metric_evaluation}
\end{table}

\vspace{-3mm}
\paragraph{Evaluation Criteria.}
We examine two measures of evaluating metrics:
\begin{inparaenum}[(1)]
    \item \textbf{Correlation}, assessed via Spearman correlation between metric-assigned and ground truth scores, and
    \item \textbf{Winrate}, the accuracy for which the metric correctly selects the summary with the higher ground truth score. For each source article, we form summary pairs and compute the average ratio of correctly ranked pairs. 
\end{inparaenum}
A desirable metric should achieve high scores for both measures, as correlation gauges true model performance whereas winrate measures the metric's accuracy in selecting the better summarizer.

\vspace{-2mm}
\subsection{Results} 
\vspace{-1mm}

We present our results in Table~\ref{tab:metric_evaluation}.
Overall, \llmcov and \alignscore serve as reliable metrics for coverage and faithfulness respectively, which we use as automatic evaluators in \S\ref{sec:results}.
Notably, metrics for coverage do not effectively measure faithfulness and vice versa, indicating that our testbed assesses these dimensions separately. 

We see that although both variants of \rouge and \bert do not achieve the highest correlation, they display moderate correlation ($0.376\sim0.527$) and winrate alignment ($0.722\sim0.815$ on average).
In contrast, we see that \bleurt and \summac exhibit poor results for both attributes.

In particular, \llmcov exhibits strong coverage performance with a Spearman correlation of $0.707$ and a winrate of $0.739$. 
For faithfulness, \llmfaith performs the best on correlation, but \alignscore, UniEval, and MiniCheck exhibit better winrates, also corroborating prior work that suggest LLMs are not yet reliable as standalone measures of faithfulness \cite{parcalabescu2024measuring, siegel2024probabilities}.
\vspace{-2mm}
\section{Method Evaluation}
\label{sec:benchmark_method_perspective_summarization}
\vspace{-2mm}

With reliable metrics established in \S\ref{sec:metric_definition_and_evaluation}, we now investigate methods for generating improved perspective summaries beyond zero-shot prompting.
Notably, due to the absence of large-scale training data, we examine several well-established methods and variants that do not rely on training data.
We use \llamathree as the default backbone for all methods.

\vspace{-3mm}
\paragraph{Prompting-Based Approaches.}
Much work on LLMs proposes inference-time methods that elicit reasoning and planning \cite{wang2023selfconsistency, press2023measuring, huang2023large, weng2023large, zhang2024selfcontrast}.
Such methods have proven effective across various tasks \citep{wang2023element, jacob2024the, saha2024branch, dhuliawala2024chain} and improve factual consistency \cite{xu2024llmrefine}.
As such, we consider two methods:
\begin{inparaenum}[(1)] 
    \item \textbf{Multi-Agent Debate} \cite{du2024improving}, where multiple LLMs iteratively update their responses based on one another, and 
    \item \textbf{Self-Refine} \cite{madaan2023selfrefine}, where an LLM iteratively critiques and revises its own output.
\end{inparaenum}
We use the default settings of three agents over three rounds for Debate and three iterations for Self-Refine.

\paragraph{Mechanistic Approach.}
A natural alternative to zero-shot inference is to direct the model's attention to salient input segments that support the overall perspective. 
Similar methods have been proposed to mitigate position biases  in LLMs using calibration-based \cite{hsieh2024found} and mechanistic approaches \cite{ratner2023parallel, hu2024explaining, liu2024lost}. 
In particular, \textbf{PINE} \cite{wang2024eliminating} modifies causal attention bidirectionally and increases the weight on specified segments.
We examine whether controlling the model's attention to segments corresponding to the desired perspective can improve coverage and faithfulness.
See \S\ref{app:experimental_setup} for additional details.

\paragraph{Reranking Generations.}
We examine a \textbf{Reranking (RR)} approach in which an untrained backbone generates multiple summaries and we select the highest-scoring summary based on \llmcov and \llmfaith.
Prior work has explored similar methods \cite{vijayakumar2018diverse, suzgun2022prompt} with notable success \cite{wei2022chain, xu2024llmrefine}. 
Benchmarking reranking examines whether the backbone is inherently capable of generating high-quality summaries.
In particular, comparing reranking with prompting-based methods, which are more commonly used to improve inference-time performance, assesses the optimal approach for perspective summarization.
For reranking-based methods, we use \qwenfourteen as the scorer backbone to avoid incorporating signals from the automatic coverage evaluation, and we generate nine summaries per input for reranking by default.

\begin{table*}[t]
    \centering
    \scriptsize
    \begin{tabularx}{\linewidth}{l|X}
        \toprule
        \bf \makecell[tl]{\vspace*{0pt}Article\\Highlights} & \hlcolor{Two years after a Ruth Bader Ginsburg's death made the elimination of that right more likely than not, President Joe Biden entered the chat\dots The fumbling about for a spell before he awakens to the fire is a recurring pattern.}{green}{30}\newline
        \hlcolor{Biden has been notably tucked away\dots With Roe, the situation is particularly galling\dots damning indictment of the administration's messaging\dots Biden appears to be trapped in a vicious cycle\dots he and his advisers appear to be gripped with anxiety that doing anything will only make things worse.}{blue}{30}\newline
        \hlcolor{Democrats need to give voters reasons to believe\dots the only way to reverse Biden's sliding popularity is for him to step forward and start providing these reasons. Our immediate goal within the Democratic Party is to "dump Biden"}{violet}{30}
        \\
        \midrule
        \midrule
        \bf Zero-Shot & The Left believes that \hlcolor{President Joe Biden's slow response to the Supreme Court's elimination of the constitutional right to an abortion and his overall lack of effective communication and decisive action are major reasons for his declining popularity and the Democrats' electoral struggles}{violet}{30}. \\
        \midrule        
        \bf DPO+RR & The Left views President \hlcolor{Joe Biden's delayed response to the SCOTUS abortion ruling, characterized by "fumbling" actions, as further exacerbating his declining public image}{green}{30} and \hlcolor{hindering his ability to provide meaningful solutions amidst various national crises}{blue}{30}, underscoring concerns that \hlcolor{his leadership style may undermine democratic values and ultimately harm Democrats' chances at re-election}{violet}{30}.\\
        \bottomrule
    \end{tabularx}
    \vspace{-2mm}
    \caption{Example summary generated by Zero-Shot and DPO+RR. Highlights indicate excerpts marked by an annotator. 
    Zero-Shot captures only one of the three excerpts, whereas DPO+RR captures all three key points.}
    \label{tab:example_summaries}
    \vspace{-4mm}
\end{table*}

\paragraph{Preference Tuning with Reranking.} 
Many studies employ reinforcement learning-based training for instruction following \cite{ouyang2022training, bai2022training, nakano2022webgpt}, with applications in summary generation \cite{stiennon2020learning, gooding2023impact, huang2024the, lee2024rlaif}.
However, these approaches typically rely on human feedback (e.g., RLHF) and labeled preference pairs (e.g., DPO \cite{rafailov2023direct}). 
Here, we examine whether preference-based training on synthetic, reranking-generated data can improve perspective summarization performance.
Namely, we consider a \textbf{DPO with Reranking (DPO+RR)} approach that iteratively repeats the procedure of generating summaries with the backbone model, scoring them with \llmcov and \llmfaith, and creating preference pairs by designating higher-scoring summaries as preferred over lower-scoring ones, which are then used to train the backbone.
We split the \polisum dataset (1816 article pairs) into train (1716) and test (100) splits to ensure that synthetic training data is generated exclusively from the train split, and repeat over 10 epochs.

\subsection{Evaluation Setup} 
\label{sec:experimental_setup}
\vspace{-1mm}

\paragraph{Automatic Evaluation.}
We automatically evaluate summary quality using two criteria. First, we assign numerical scores to summaries using \llmcov and \alignscore (cf.~\S\ref{sec:metric_definition_and_evaluation}). Second, we compute instance-level rankings across all test articles to assess relative method performance. 
However, as automatic metrics do not rank methods perfectly (cf.~Table~\ref{tab:metric_evaluation}), we address this by fitting a Bradley-Terry model to the pairwise comparisons derived from raw scores and performing bootstrap resampling over the test documents to obtain 95\% confidence intervals.
This avoids naive "rank-then-average" methods that can yield cyclic or inconsistent preferences when pairwise comparisons do not form a strict total ordering. 
We use the split test set of 200 input documents for automatic evaluation.
Refer to \S\ref{app:ranking_details} for details on the ranking procedure.

\vspace{-2mm}
\paragraph{Human Evaluation.} 
\label{sec:human_evaluation}

We collect human judgments on summary quality by having annotators review input documents and their corresponding model-generated summaries.
Analogous to the procedure in \S\ref{sec:assessing_metric_quality}, annotators first extract key points from both documents and summaries, then identify which document key points each summary includes or omits, and which summary key points appear in the document. 
This process yields coverage and faithfulness scores computed as in Eqs.~\eqref{eqn:coverage} and \eqref{eqn:faithfulness}.
See \S\ref{app:annotation_details} for further annotation details.
\begin{figure*}[t]
    \centering
    \begin{subfigure}[t]{0.67\linewidth}
        \centering
        \includegraphics[width=\linewidth]{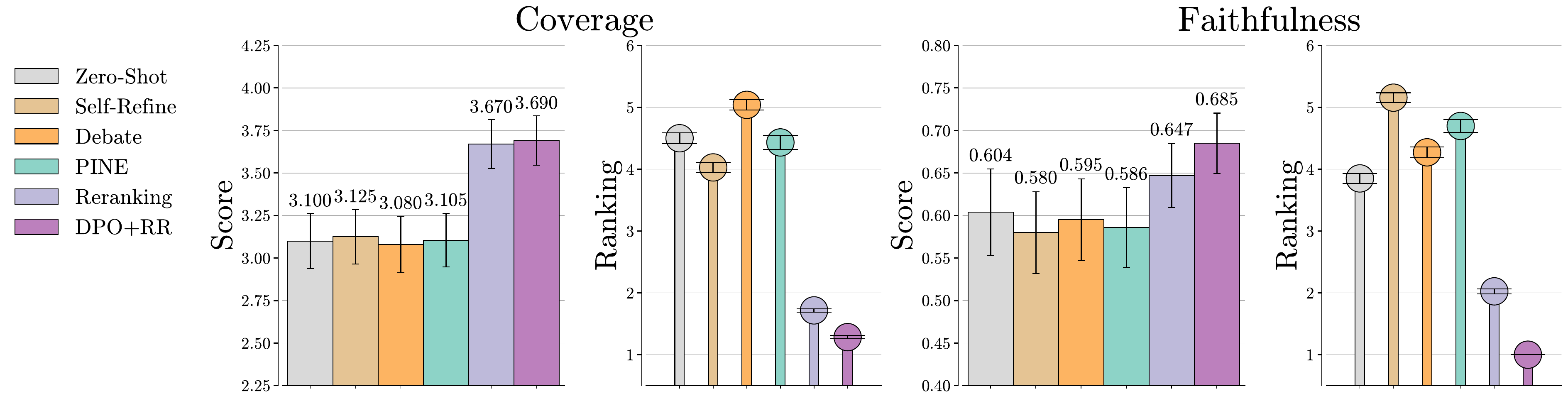}
        \caption{Automatic evaluation results. Higher values indicate better performance for Score (Bars), while lower values are better for Ranking (Lolipops). 
        Coverage scores range from 1 to 5, while faithfulness scores lie in the interval $[0,1]$.
        DPO+RR achieves the highest scores and best average rank, followed by Reranking. 
        Other methods show similar performance in both coverage and faithfulness.}
        \label{fig:summary_metric_evaluation}
        \vspace{-2mm}
    \end{subfigure}
    \hspace{3mm}
    \begin{subfigure}[t]{0.3\linewidth}
        \centering
        \includegraphics[width=\linewidth]{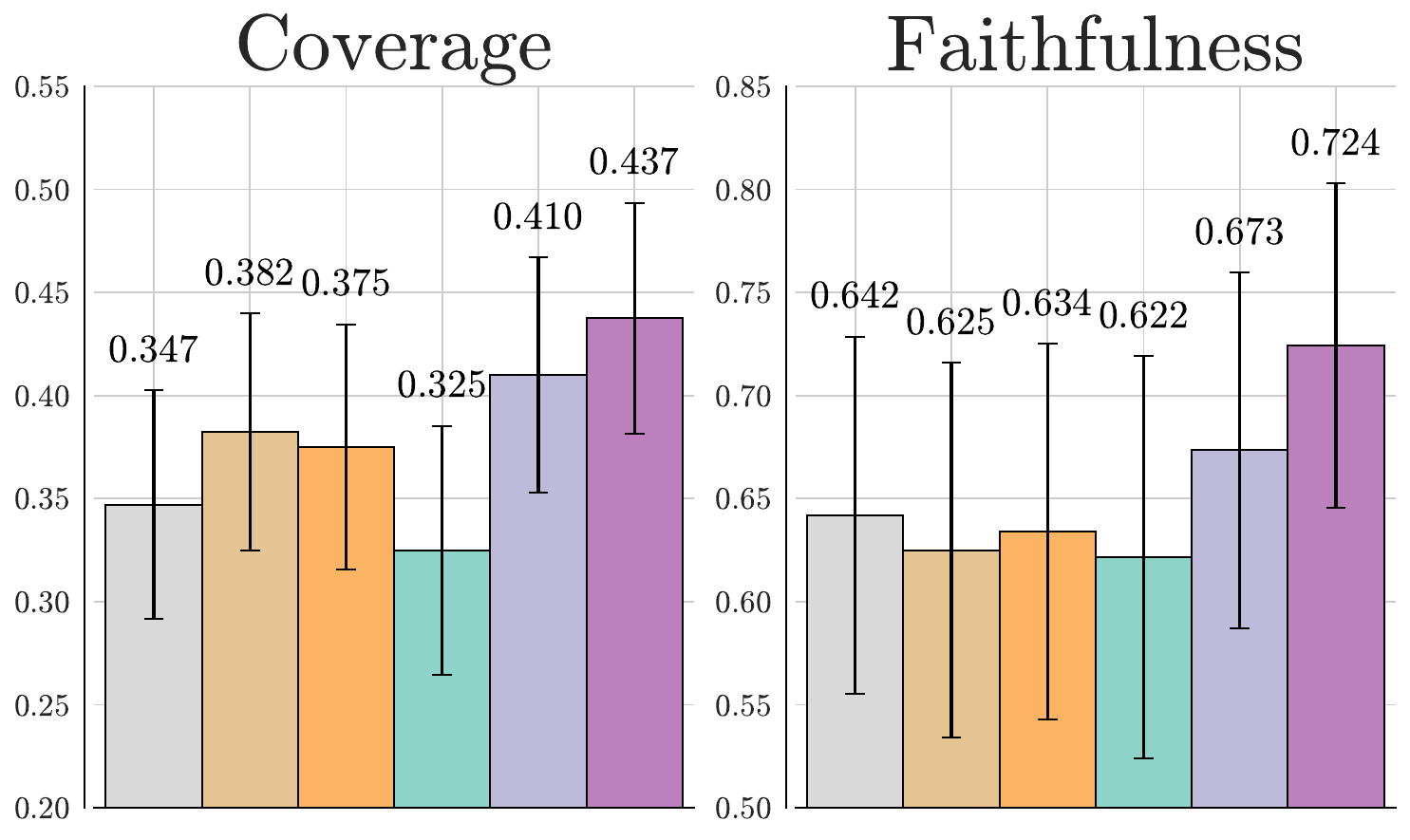}
        \caption{Human evaluation results. 
        Higher is better for both attributes.
        DPO+RR achieves the best performance for both attributes, followed by Reranking. 
        Scores lie in $[0,1]$ for both attributes.}
        \label{fig:summary_human_evaluation}
        \vspace{-2mm}
    \end{subfigure}
    \caption{Automatic (left) and human (right) evaluation results. 
    For clarity, note that $y$-axes do not begin at 0 in score plots.
    Reranking-based methods perform best across both evaluation regimes, with DPO+RR achieving the highest overall performance in both coverage and faithfulness.
    Error bars represent 95\% confidence intervals (CI).}
    \label{fig:combined_summary_evaluation}
    \vspace{-4mm}
\end{figure*}

\vspace{-2mm}
\section{Results}
\label{sec:results}
\vspace{-2mm}

We present coverage and faithfulness results in Figures~\ref{fig:summary_metric_evaluation} (automatic evaluation) and \ref{fig:summary_human_evaluation} (human evaluation), and provide generated example summaries in 
Table~\ref{tab:example_summaries}. 
We include additional examples in \S\ref{app:example_summaries}.

\vspace{-2mm}
\subsection{Automatic Evaluation}
We observe that DPO+RR achieves the highest performance on both metrics, improving coverage and faithfulness scores by $0.590$ and $0.081$, corresponding to approximately 12\% and 8\% gains, respectively. 
Reranking is a strong baseline, outperforming all other methods by considerable margins, corroborating prior work on the benefits of re-ranking (e.g., \cite{horvitz2024tinystyler}).
In contrast, zero-shot inference, prompting methods, and PINE show minimal score differences. 
Although Self-Refine marginally improves coverage over zero-shot inference, all methods except reranking yield lower faithfulness scores.

\vspace{-3mm}
\subsection{Human Evaluation}
\label{sec:results_human_evaluation}
\vspace{-1mm}

DPO+RR achieves the highest human evaluation scores ($0.437$ for coverage and $0.724$ for faithfulness), with Reranking close behind ($0.410$ and $ 0.673$, respectively). 
Prompting-based methods improve coverage over zero-shot inference ($0.347$ for coverage and $0.642$ for faithfulness) but yield similar faithfulness scores. PINE shows no performance gains for either attribute.

\begin{table}[t]
    \centering
    \begin{tabular}{l|cc}
    \toprule
    \bf Ratio      & \bf Document & \bf Summary\\  
    \midrule
    \midrule 
    $R(\cdot\mid \cdot)$   & $0.672 \pm 0.262$ & $0.918 \pm 0.173$ \\
    Random  & $0.235 \pm 0.322$ & $0.650 \pm 0.380$ \\
    \bottomrule
    \end{tabular}
    \vspace{-3mm}
    \caption{Inter-Annotator Agreement (IAA) results. 
    Values lie between the interval $[0,1]$. 
    We observe substantial agreement for both document- and summary-level key point extraction.}
    \vspace{-5mm}
    \label{tab:iaa}
\end{table}

\begin{table}[t]
    \centering
    \resizebox{\columnwidth}{!}{
    \begin{tabular}{l|ccc}
    \toprule
    \bf Method        & $|K_D\cap K_S|$ & $|K_D \setminus K_S|$  & $|K_S \setminus K_D|$\\
    \midrule
    \midrule 
    Zero-Shot          & \cellcolor{orange!40} $1.338 \pm 0.894$ & \cellcolor{orange!30} $3.059 \pm 1.254$ & \cellcolor{orange!20} $0.765 \pm 0.855$ \\
    Self-Refine       & \cellcolor{orange!50} $1.412 \pm 1.097$ & \cellcolor{orange!40} $2.912 \pm 1.288$ & \cellcolor{orange!10} $0.794 \pm 0.729$ \\
    Debate            & \cellcolor{orange!45} $1.368 \pm 0.847$ & \cellcolor{orange!35} $2.971 \pm 1.291$ & \cellcolor{orange!30} $0.735 \pm 0.790$ \\
    PINE              & \cellcolor{orange!30} $1.206 \pm 0.854$ & \cellcolor{orange!25} $3.235 \pm 1.350$ & \cellcolor{orange!40} $0.706 \pm 0.799$ \\
    Reranking         & \cellcolor{orange!55} $1.544 \pm 0.916$ & \cellcolor{orange!45} $2.882 \pm 1.320$ & \cellcolor{orange!30} $0.735 \pm 0.828$ \\
    DPO+RR               & \cellcolor{orange!70} $1.721 \pm 0.889$ & \cellcolor{orange!70} $2.500 \pm 1.080$ & \cellcolor{orange!70} $0.618 \pm 0.739$ \\
    \bottomrule
    \end{tabular}
    }
    \vspace{-3mm}
    \caption{Statistics for key point inclusion for each method with standard deviation. 
    $|K_D \cap K_S|$, $|K_D \setminus K_S|$, and $|K_S \setminus K_D|$ denote the average number of key points included, omitted, and hallucinated, respectively.}
    \label{tab:key_point_inclusion}
    \vspace{-5mm}
\end{table}

\vspace{-2mm}
\paragraph{Inter-Annotator Agreement (IAA).}  
We measure IAA by counting the number of excerpts with non-trivial overlap between annotators. 
Formally, given excerpts from two annotators $A$ and $B$ (from a document or a summary), denoted as $E_A = \{e_1^{A}, e_2^{A}, \dots\}$ and $E_B = \{e_1^{B}, e_2^{B}, \dots\}$, we define a matching function $M(E_A, E_B)$ that counts the number of strings in $E_A$ matched to at most one string in $E_B$. 
We then compute $R(A \mid B) = {|M(E_A, E_B)|}/{|E_A|}$ and $R(B \mid A) = {|M(E_A, E_B)|}/{|E_B|}$, and take their average to obtain the overall annotator overlap \(R(\cdot \mid \cdot)\).
To assess overlap, we provide overlapping annotations to pairs of annotators across five documents and evaluate agreement for both document-level and summary-level key point extraction. 
Additionally, we establish a random baseline for annotator overlap by sampling highlight counts and lengths for documents and summaries that match the observed mean and variance in the real annotations.
Further details are provided in \S\ref{app:iaa_details}.

Results are presented in Table~\ref{tab:iaa}. 
Observe that annotators exhibit substantial overlap in both document- and summary-level annotations that considerably exceed the random baseline.
We also see higher agreement for summaries than for documents, which we attribute to summaries being more concise and explicitly including key points.

Overall, our results show that while prompting-based and attention modification methods offer little improvement over zero-shot prompting, reranking-based methods significantly improves coverage and faithfulness. 
In particular, employing DPO-based training further boosts faithfulness, even when using self-generated synthetic data.
\section{Analysis}
\label{sec:analysis}
\vspace{-2mm}

\subsection{Summary Characteristics}
\label{sec:summary_characteristics}
\vspace{-1mm}

Here, we examine the summaries generated by each method and assess their key point inclusion patterns, abstractiveness, and length.

\paragraph{Key Point Inclusion.}  
Beyond coverage and faithfulness, we evaluate how each method incorporates key points. 
For an article $D$ with key points $K_D$ and a summary $S$ with key points $K_S$, we compute the average number of key points included ($|K_D \cap K_S|$), omitted ($|K_D \setminus K_S|$), and hallucinated ($|K_S \setminus K_D|$). 

Results are shown in Table~\ref{tab:key_point_inclusion}. 
We observe that DPO+RR includes more relevant key points while minimizing hallucinations and omissions compared to other methods. 
In contrast, PINE is more conservative, reducing hallucinations but omitting more key points. 
Self-Refine retains additional key points yet introduces more hallucinations, while Debate shows only slight improvements over the zero-shot baseline.

\begin{table}[t]
    \centering
    \resizebox{\columnwidth}{!}{
    \begin{tabular}{l|cc}
    \toprule
    \bf Method        & \bf Novel $4$-gram ($\uparrow$) & \bf EF Density ($\downarrow$) \\
    \midrule\midrule 
    Zero-Shot          & \cellcolor{orange!62} $0.930 \pm 0.104$ & \cellcolor{orange!47} $1.815 \pm 1.614$ \\
    Self-Refine        & \cellcolor{orange!74} $0.946 \pm 0.094$ & \cellcolor{orange!22} $1.470 \pm 1.307$ \\
    Debate             & \cellcolor{orange!80} $0.954 \pm 0.088$ & \cellcolor{orange!26} $1.571 \pm 1.162$ \\
    PINE               & \cellcolor{orange!50} $0.848 \pm 0.074$ & \cellcolor{orange!70} $3.340 \pm 4.801$ \\
    Reranking          & \cellcolor{orange!76} $0.949 \pm 0.217$ & \cellcolor{orange!21} $1.445 \pm 0.914$ \\
    DPO+RR             & \cellcolor{orange!79} $0.953 \pm 0.079$ & \cellcolor{orange!20} $1.415 \pm 1.039$ \\
    \bottomrule
    \end{tabular}
    }
    \vspace{-3mm}
    \caption{Abstractiveness statistics for each method, measured by novel $n$-gram ratios and extractive fragment density.
    Arrows indicate higher abstractiveness.}
    \vspace{-5mm}
    \label{tab:summary_abstractiveness}
\end{table}

\begin{figure*}[t]
    \centering
    \begin{subfigure}[t]{0.32\textwidth}
        \centering
        \includegraphics[width=\linewidth]{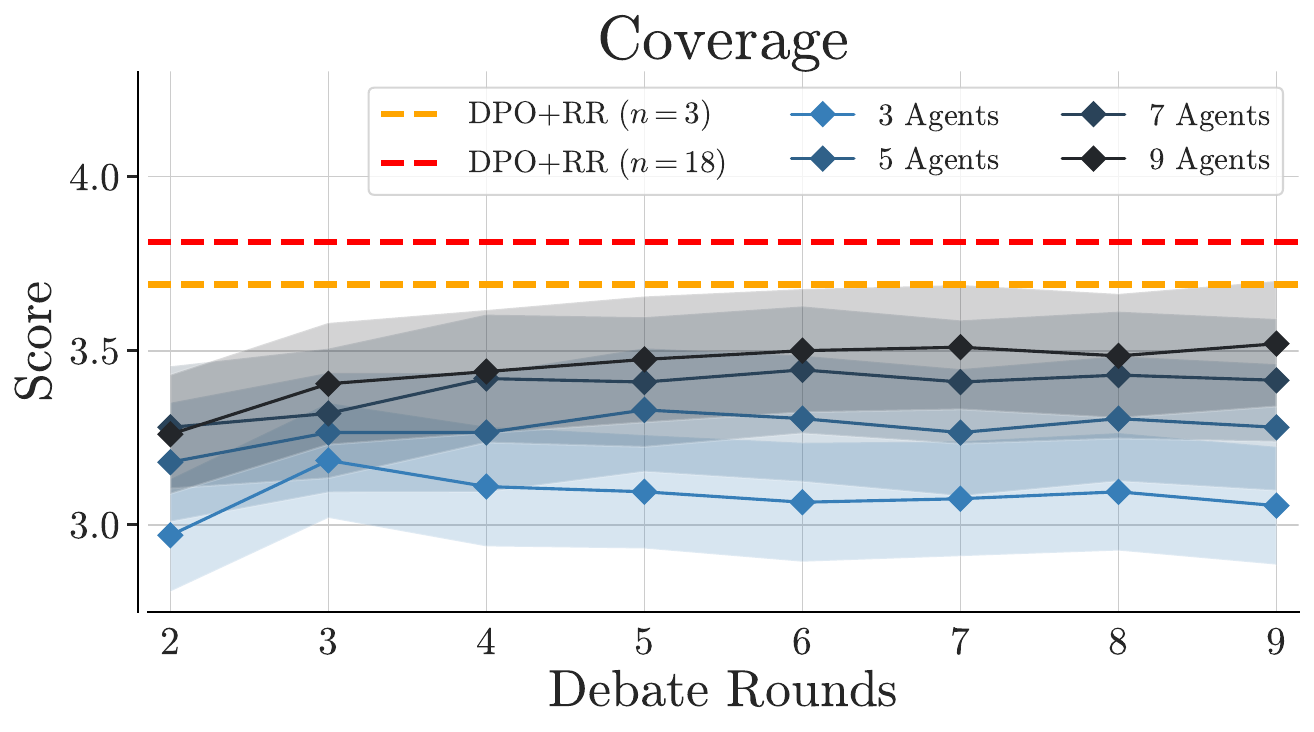}
        \hfill
        \includegraphics[width=\linewidth]{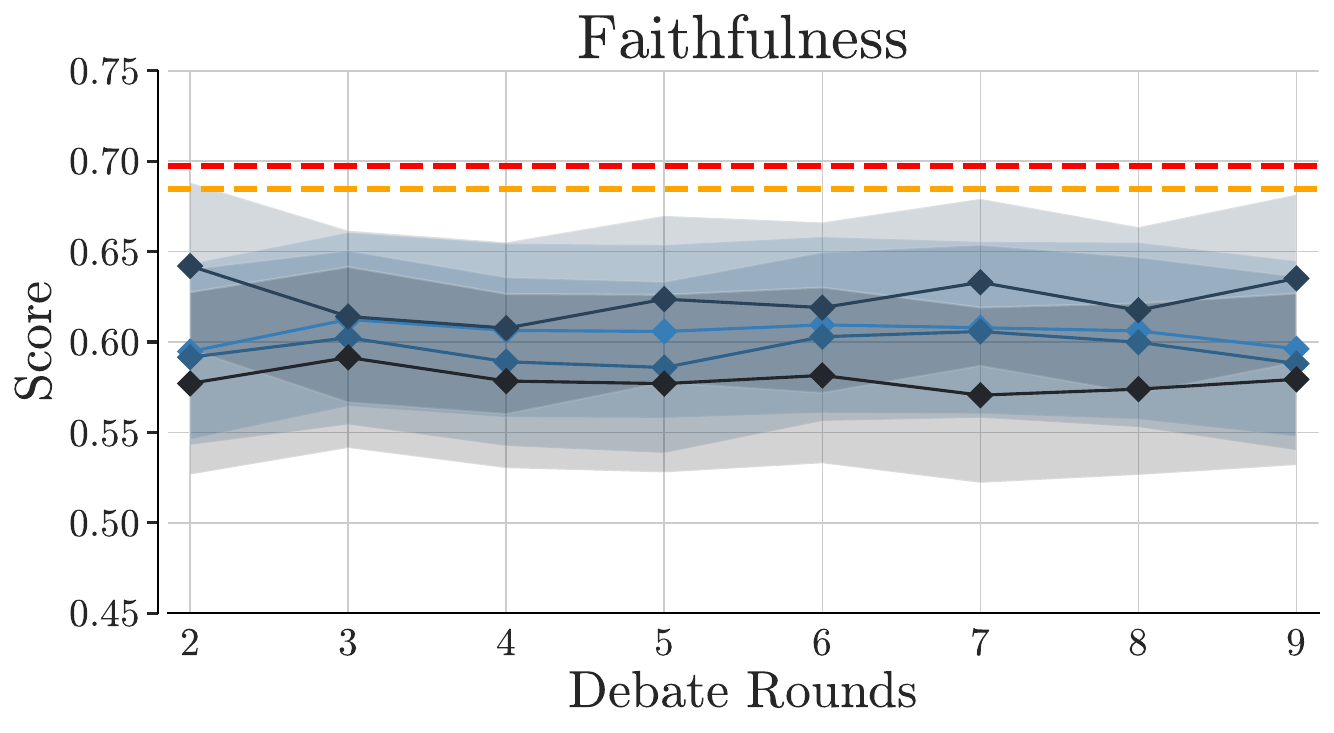}
        \caption{Debate performance across varying agent counts and debate rounds.}
        \label{fig:ablation_debate}
    \end{subfigure}
    \hfill
    \begin{subfigure}[t]{0.32\textwidth}
        \centering
        \includegraphics[width=\linewidth]{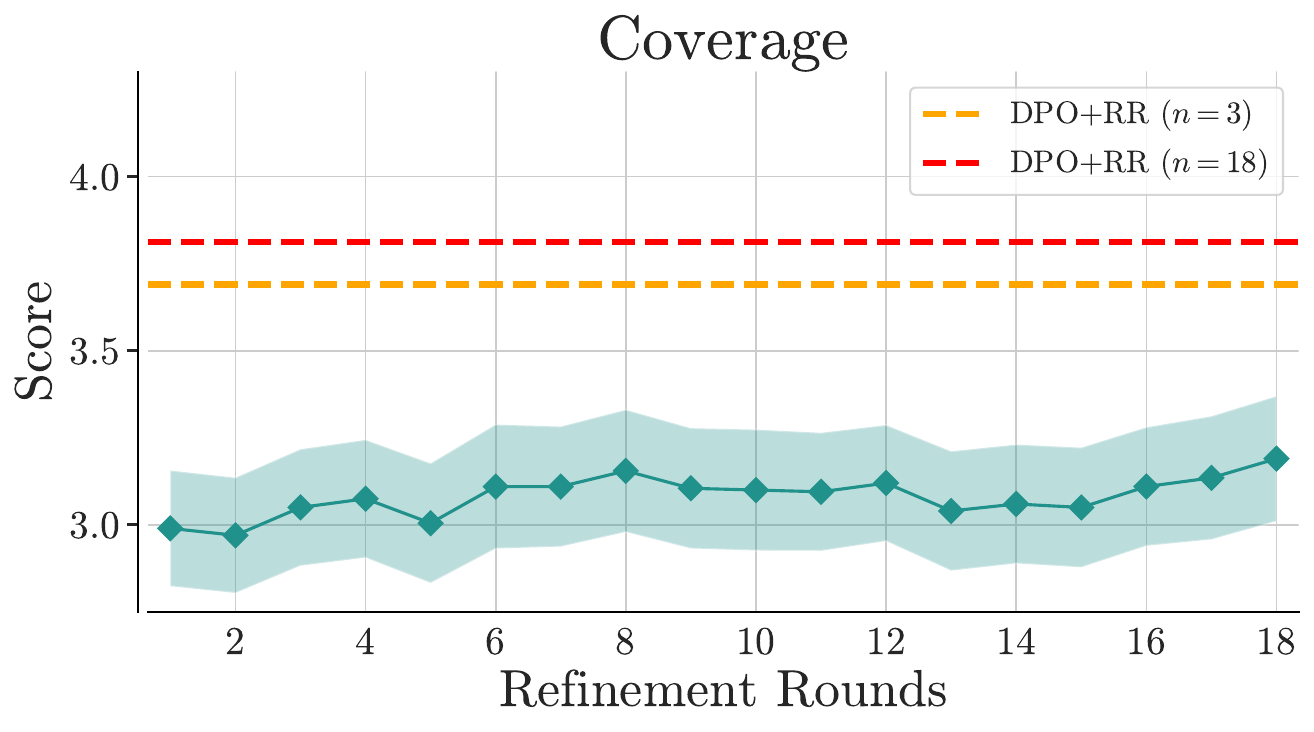}
        \includegraphics[width=\linewidth]{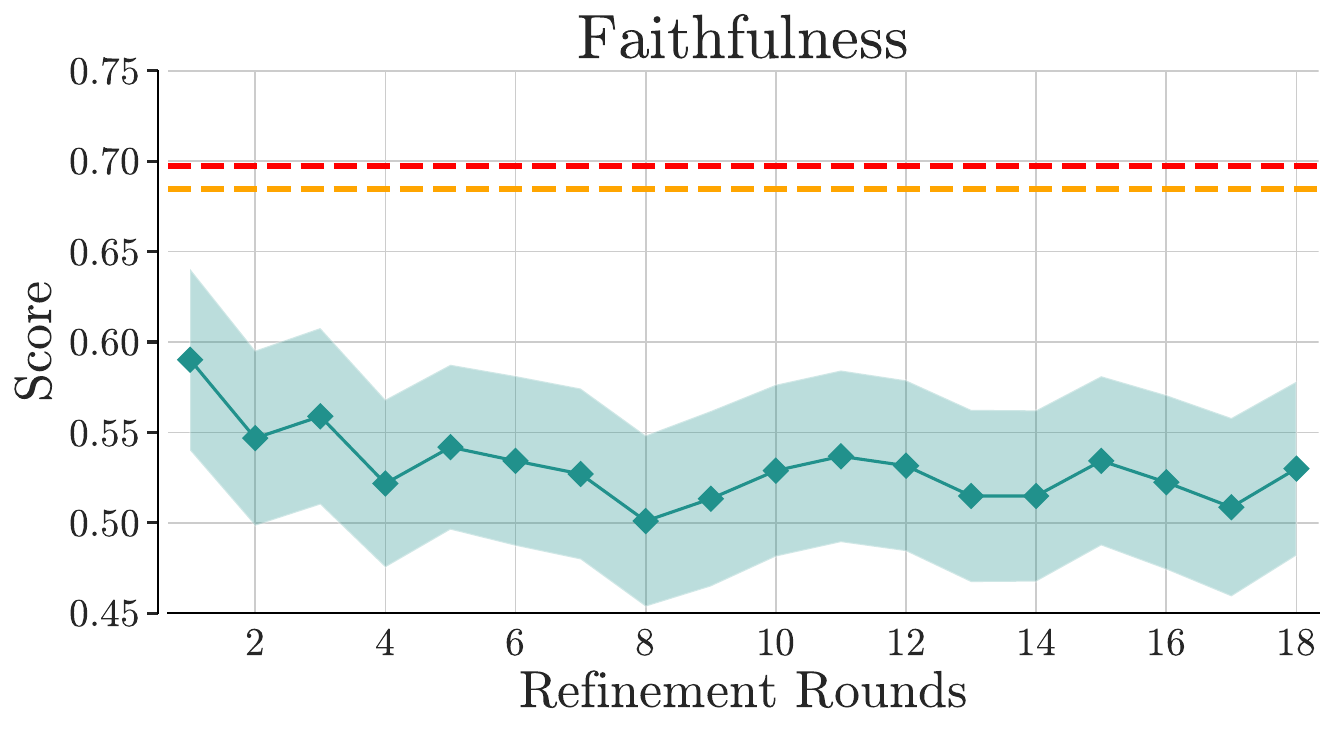}   
        \caption{Self-Refine performance across varying refinement rounds.}
        \label{fig:ablation_self_refine}
    \end{subfigure}
    \hfill
    \begin{subfigure}[t]{0.32\textwidth}
        \centering
        \includegraphics[width=\linewidth]{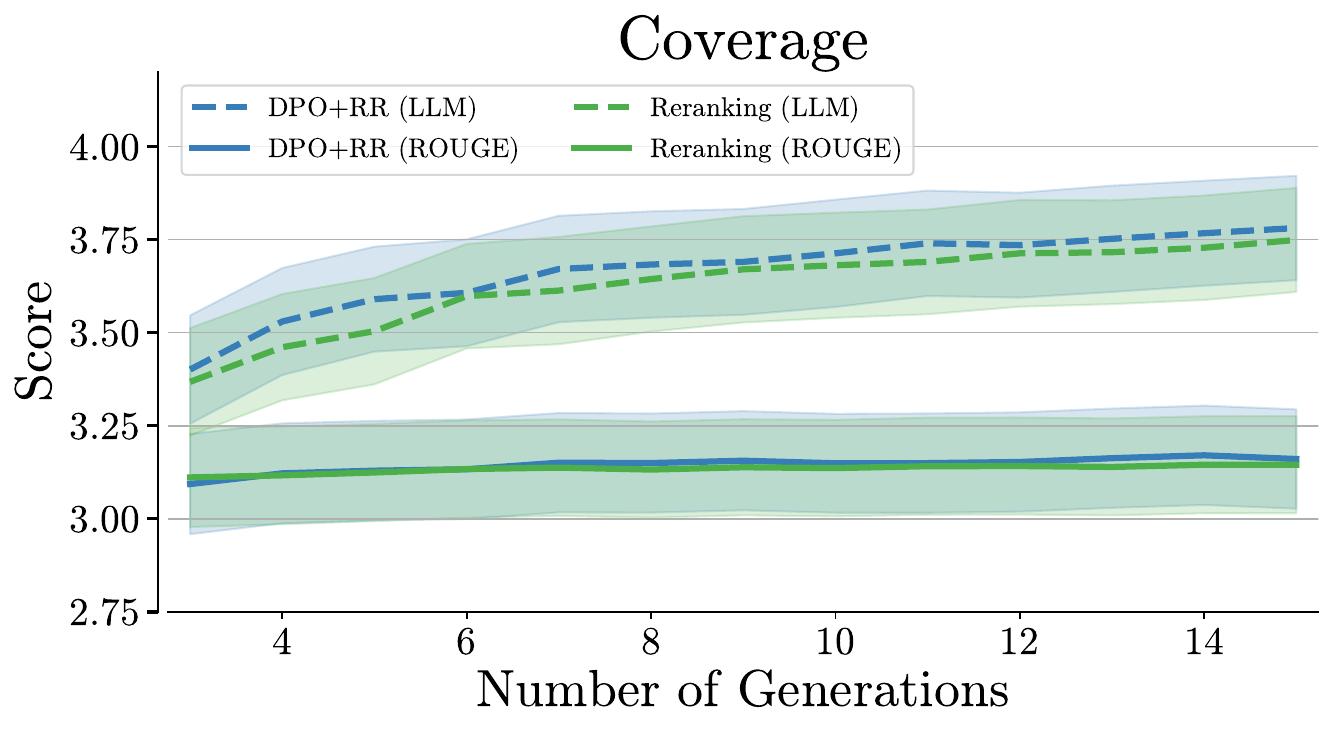}
        \hfill
        \includegraphics[width=\linewidth]{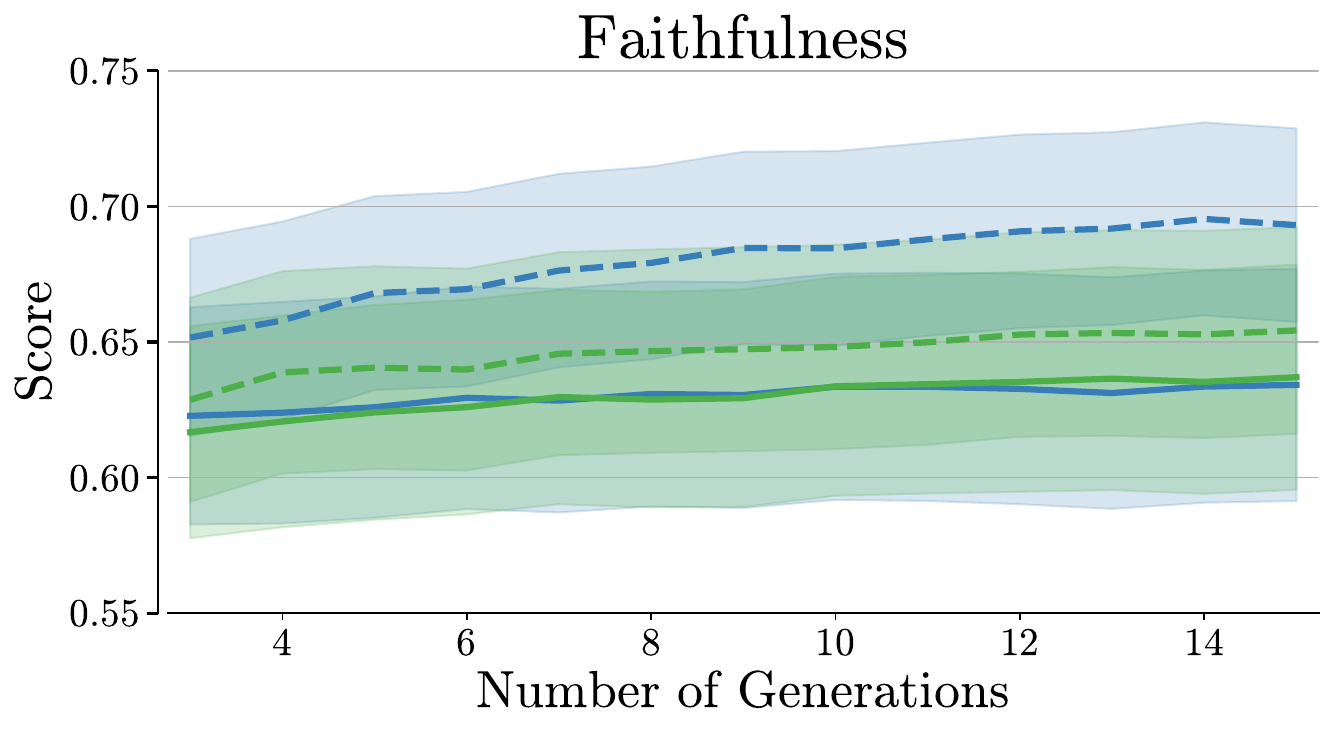}
        \caption{Comparison of LLM- and \rouge-based proxies for reranking methods.}
        \label{fig:ablation_reranking}
    \end{subfigure}    
    \caption{Ablation study results. 
    Figures~\ref{fig:ablation_debate} and \ref{fig:ablation_self_refine} show that both prompting-based methods consistently underperform compared to reranking-based methods across all resource settings. 
    Figure~\ref{fig:ablation_reranking} shows that using a \rouge-based proxy metric yields worse performance than LLM-based proxy metrics.}
    \label{fig:ablation}
    \vspace{-2mm}
\end{figure*}

\vspace{-2mm}
\paragraph{Summary Abstractiveness.}
We assess abstractiveness using two metrics:
\begin{inparaenum}[(1)] 
    \item \emph{Novel $n$-gram ratios} \cite{see2017get}, which measure the proportion of $n$-grams in the summary absent from the source (with $n=4$), and 
    \item \emph{Extractive fragment density} \cite{grusky2018newsroom}, which quantifies the continuity of extracted spans. 
\end{inparaenum}
Higher novel $n$-gram ratios and lower extractive fragment density indicate greater abstractiveness. 
We include additional results for analysis in \S\ref{app:methods_supplementary_analysis}.

Table~\ref{tab:summary_abstractiveness} shows our results. 
Notably, PINE exhibits lower novel $n$-gram ratios and higher extractive fragment density than other methods, indicating that PINE favors more extractive summaries.
In contrast, DPO+RR yields higher abstractiveness than zero-shot inference while also improving faithfulness (cf.~\S\ref{sec:results}). 
This suggests that DPO+RR not only encourages extraction of source content but also generates summaries with more novel tokens.

\vspace{-3mm}
\subsection{Ablation Studies}
\label{sec:ablation_studies}
\vspace{-1mm}

We conduct ablation studies to determine whether prompting-based methods outperform Reranking and DPO+RR under more resourceful generation settings, as measured by automated metrics.
See Figure~\ref{fig:ablation} for all results.

\vspace{-2mm}
\paragraph{Debate: Agents and Rounds.}
We vary the number of rounds $n \in \{2,3,\dots,9\}$ and agents $m \in \{3,5,7,9\}$ in Multi-Agent Debate. 
For reference, we report results for DPO+RR in two settings: generating 3 (base setting) and generating 18 summaries (approximate upper bound).

From Figure~\ref{fig:ablation_debate}, we observe that increasing the number of agents improves coverage but not faithfulness.
With $m=9$ agents, Debate slightly outperforms DPO+RR with 3 reranked generations for $n\geq4$ in coverage but falls short of DPO+RR with 18 generations. 
For faithfulness, Debate remains consistently below DPO+RR in all settings.

\vspace{-2mm}
\paragraph{Self-Refine: Refinement Rounds.}  
We evaluate Self-Refine over various numbers of refinement rounds ($n \in \{2,3,\dots,18\}$).
Results are shown in Figure~\ref{fig:ablation_self_refine}. 
We observe that coverage improves with more rounds, whereas faithfulness does not. 
Nevertheless, Self-Refine underperforms DPO+RR in both settings across all rounds.

\vspace{-2mm}
\paragraph{Reranking: \rouge as Proxy Metric.}  
To assess the effectiveness of LLM-based proxy metrics, we compare Reranking and DPO+RR with variants that use \rouge as the proxy. 
Following the training procedure in \S\ref{sec:benchmark_method_perspective_summarization}, we replace the original proxy with the average \rougen{n} score (for $n \in \{1, 2, L\}$) computed across both precision and recall. 
As shown in Figure~\ref{fig:ablation_reranking}, the \rouge-based variant underperforms across all settings.
\vspace{-3mm}
\section{Conclusion}
\label{sec:conclusion_discussion}
\vspace{-3mm}

In this paper, we identify reliable evaluation metrics for measuring perspective summary quality and investigate LLM-based methods for generating improved summaries beyond zero-shot inference. 
We construct a test dataset using human annotations to benchmark existing summarization metrics for coverage and faithfulness. 
We find that traditional metrics such as \rouge and \bert underperform, while language model–based metrics such as \alignscore and prompting-based scoring serve as strong evaluators. 
Using these metrics, we show that reranking-based methods outperform prompting frameworks and significantly improve performance over zero-shot inference. 
Moreover, preference tuning with self-generated, reranking-labeled data further boosts performance, particularly in terms of faithfulness. 
We recommend that future work examine the transferability of our findings to domains beyond political perspectives and whether similar improvements can be achieved in other multi-document summarization tasks.

\section*{Limitations}

We acknowledge two limitations in our work. 
First, we focus on evaluating existing summarization metrics commonly used in the literature and benchmark those applied to perspective summarization.
As we show that existing metrics achieve satisfactory accuracy for evaluating perspective summaries, we do not investigate the development of a novel metric tailored specifically for measuring coverage and faithfulness in this setting.
We leave this as a promising direction for future work.
Second, we primarily investigated methods for perspective summary generation that do not rely on human-labeled training data, given the infeasibility of collecting such data. 
Although our experiments with preference tuning using synthetically generated data show performance improvements, future studies should examine the benefits of human-curated training data.

\section*{Ethical Considerations}
In this paper, we focus on metrics to accurately measure the unbiasedness of perspective summaries through the attributes of coverage and faithfulness, and we show that certain methods yield higher performance on these attributes. 
Our work aims to ensure fair representation and reduce hallucinations in opinion-based summarization. 
While it is unclear whether these findings could be misused to generate more biased summaries, we acknowledge that such risks are not negligible.

\section*{Acknowledgements}
This work was supported in part by the Knight First Amendment Institute at Columbia University, National Science Foundation Graduate Research Fellowship DGE-2036197, the Columbia University Provost Diversity Fellowship, and the Columbia School of Engineering and Applied Sciences Presidential Fellowship.
Any opinion, findings, and conclusions or recommendations expressed in this material are those of the authors and do not necessarily reflect the views of the Knight First Amendment Institute or National Science Foundation.
We thank the anonymous reviewers for providing feedback on an earlier draft of the work.

\bibliography{custom}

\begin{thebibliography}{79}
\providecommand{\natexlab}[1]{#1}

\bibitem[{Amplayo et~al.(2021)Amplayo, Angelidis, and Lapata}]{amplayo2021aspect}
Reinald~Kim Amplayo, Stefanos Angelidis, and Mirella Lapata. 2021.
\newblock \href {https://doi.org/10.18653/v1/2021.emnlp-main.528} {Aspect-controllable opinion summarization}.
\newblock In \emph{Proceedings of the 2021 Conference on Empirical Methods in Natural Language Processing}, pages 6578--6593, Online and Punta Cana, Dominican Republic. Association for Computational Linguistics.

\bibitem[{Bai et~al.(2022)Bai, Jones, Ndousse, Askell, Chen, DasSarma, Drain, Fort, Ganguli, Henighan, Joseph, Kadavath, Kernion, Conerly, El-Showk, Elhage, Hatfield-Dodds, Hernandez, Hume, Johnston, Kravec, Lovitt, Nanda, Olsson, Amodei, Brown, Clark, McCandlish, Olah, Mann, and Kaplan}]{bai2022training}
Yuntao Bai, Andy Jones, Kamal Ndousse, Amanda Askell, Anna Chen, Nova DasSarma, Dawn Drain, Stanislav Fort, Deep Ganguli, Tom Henighan, Nicholas Joseph, Saurav Kadavath, Jackson Kernion, Tom Conerly, Sheer El-Showk, Nelson Elhage, Zac Hatfield-Dodds, Danny Hernandez, Tristan Hume, Scott Johnston, Shauna Kravec, Liane Lovitt, Neel Nanda, Catherine Olsson, Dario Amodei, Tom Brown, Jack Clark, Sam McCandlish, Chris Olah, Ben Mann, and Jared Kaplan. 2022.
\newblock \href {https://arxiv.org/abs/2204.05862} {Training a helpful and harmless assistant with reinforcement learning from human feedback}.
\newblock \emph{Preprint}, arXiv:2204.05862.

\bibitem[{Bakker et~al.(2022)Bakker, Chadwick, Sheahan, Tessler, Campbell-Gillingham, Balaguer, McAleese, Glaese, Aslanides, Botvinick, and Summerfield}]{bakker2022finetuning}
Michiel Bakker, Martin Chadwick, Hannah Sheahan, Michael Tessler, Lucy Campbell-Gillingham, Jan Balaguer, Nat McAleese, Amelia Glaese, John Aslanides, Matt Botvinick, and Christopher Summerfield. 2022.
\newblock \href {https://proceedings.neurips.cc/paper_files/paper/2022/file/f978c8f3b5f399cae464e85f72e28503-Paper-Conference.pdf} {Fine-tuning language models to find agreement among humans with diverse preferences}.
\newblock In \emph{Advances in Neural Information Processing Systems}, volume~35, pages 38176--38189. Curran Associates, Inc.

\bibitem[{Banerjee and Lavie(2005)}]{banerjee-meteor}
Satanjeev Banerjee and Alon Lavie. 2005.
\newblock \href {https://aclanthology.org/W05-0909/} {{METEOR}: An automatic metric for {MT} evaluation with improved correlation with human judgments}.
\newblock In \emph{Proceedings of the {ACL} Workshop on Intrinsic and Extrinsic Evaluation Measures for Machine Translation and/or Summarization}, pages 65--72, Ann Arbor, Michigan. Association for Computational Linguistics.

\bibitem[{Bradley and Terry(1952)}]{bradley1952rank}
Ralph~Allan Bradley and Milton~E. Terry. 1952.
\newblock \href {https://api.semanticscholar.org/CorpusID:125209808} {Rank analysis of incomplete block designs: I. the method of paired comparisons}.
\newblock \emph{Biometrika}, 39:324.

\bibitem[{Bra{\v{z}}inskas et~al.(2020)Bra{\v{z}}inskas, Lapata, and Titov}]{brazinskas-opinion}
Arthur Bra{\v{z}}inskas, Mirella Lapata, and Ivan Titov. 2020.
\newblock \href {https://doi.org/10.18653/v1/2020.acl-main.461} {Unsupervised opinion summarization as copycat-review generation}.
\newblock In \emph{Proceedings of the 58th Annual Meeting of the Association for Computational Linguistics}, pages 5151--5169, Online. Association for Computational Linguistics.

\bibitem[{Chen et~al.(2022)Chen, Li, Gao, and Zhang}]{chen-multi}
Xiuying Chen, Mingzhe Li, Xin Gao, and Xiangliang Zhang. 2022.
\newblock \href {https://proceedings.neurips.cc/paper_files/paper/2022/file/9b6d7202750e8e32cd5270eb7fc131f7-Paper-Conference.pdf} {Towards improving faithfulness in abstractive summarization}.
\newblock In \emph{Advances in Neural Information Processing Systems}, volume~35, pages 24516--24528. Curran Associates, Inc.

\bibitem[{Chen et~al.(2023)Chen, Wang, Jiang, Shi, and Xu}]{chen2023exploring}
Yi~Chen, Rui Wang, Haiyun Jiang, Shuming Shi, and Ruifeng Xu. 2023.
\newblock \href {https://doi.org/10.18653/v1/2023.findings-ijcnlp.32} {Exploring the use of large language models for reference-free text quality evaluation: An empirical study}.
\newblock In \emph{Findings of the Association for Computational Linguistics: IJCNLP-AACL 2023 (Findings)}, pages 361--374, Nusa Dua, Bali. Association for Computational Linguistics.

\bibitem[{Chhabra et~al.(2024)Chhabra, Askari, and Mohapatra}]{chhabra2024revisiting}
Anshuman Chhabra, Hadi Askari, and Prasant Mohapatra. 2024.
\newblock \href {https://arxiv.org/abs/2401.01989} {Revisiting zero-shot abstractive summarization in the era of large language models from the perspective of position bias}.
\newblock \emph{Preprint}, arXiv:2401.01989.

\bibitem[{Chiang and Lee(2023)}]{chiang2023large}
Cheng-Han Chiang and Hung-yi Lee. 2023.
\newblock \href {https://doi.org/10.18653/v1/2023.acl-long.870} {Can large language models be an alternative to human evaluations?}
\newblock In \emph{Proceedings of the 61st Annual Meeting of the Association for Computational Linguistics (Volume 1: Long Papers)}, pages 15607--15631, Toronto, Canada. Association for Computational Linguistics.

\bibitem[{Dao(2023)}]{dao2023flashattention2}
Tri Dao. 2023.
\newblock \href {https://arxiv.org/abs/2307.08691} {Flashattention-2: Faster attention with better parallelism and work partitioning}.
\newblock \emph{Preprint}, arXiv:2307.08691.

\bibitem[{Deas and McKeown(2025)}]{deas2025summarization}
Nicholas Deas and Kathleen McKeown. 2025.
\newblock \href {https://aclanthology.org/2025.coling-main.539/} {Summarization of opinionated political documents with varied perspectives}.
\newblock In \emph{Proceedings of the 31st International Conference on Computational Linguistics}, pages 8088--8108, Abu Dhabi, UAE. Association for Computational Linguistics.

\bibitem[{Dhuliawala et~al.(2024)Dhuliawala, Komeili, Xu, Raileanu, Li, Celikyilmaz, and Weston}]{dhuliawala2024chain}
Shehzaad Dhuliawala, Mojtaba Komeili, Jing Xu, Roberta Raileanu, Xian Li, Asli Celikyilmaz, and Jason Weston. 2024.
\newblock \href {https://doi.org/10.18653/v1/2024.findings-acl.212} {Chain-of-verification reduces hallucination in large language models}.
\newblock In \emph{Findings of the Association for Computational Linguistics: ACL 2024}, pages 3563--3578, Bangkok, Thailand. Association for Computational Linguistics.

\bibitem[{Dreyer et~al.(2023)Dreyer, Liu, Nan, Atluri, and Ravi}]{dreyer-tradeoff}
Markus Dreyer, Mengwen Liu, Feng Nan, Sandeep Atluri, and Sujith Ravi. 2023.
\newblock \href {https://doi.org/10.18653/v1/2023.findings-eacl.156} {Evaluating the tradeoff between abstractiveness and factuality in abstractive summarization}.
\newblock In \emph{Findings of the Association for Computational Linguistics: EACL 2023}, pages 2089--2105, Dubrovnik, Croatia. Association for Computational Linguistics.

\bibitem[{Du et~al.(2024)Du, Li, Torralba, Tenenbaum, and Mordatch}]{du2024improving}
Yilun Du, Shuang Li, Antonio Torralba, Joshua~B. Tenenbaum, and Igor Mordatch. 2024.
\newblock Improving factuality and reasoning in language models through multiagent debate.
\newblock In \emph{Proceedings of the 41st International Conference on Machine Learning}, ICML'24. JMLR.org.

\bibitem[{Dubois et~al.(2023)Dubois, Li, Taori, Zhang, Gulrajani, Ba, Guestrin, Liang, and Hashimoto}]{dubois2023alpacafarm}
Yann Dubois, Xuechen Li, Rohan Taori, Tianyi Zhang, Ishaan Gulrajani, Jimmy Ba, Carlos Guestrin, Percy Liang, and Tatsunori Hashimoto. 2023.
\newblock \href {https://openreview.net/forum?id=4hturzLcKX} {Alpacafarm: A simulation framework for methods that learn from human feedback}.
\newblock In \emph{Thirty-seventh Conference on Neural Information Processing Systems}.

\bibitem[{Durmus et~al.(2020)Durmus, He, and Diab}]{durmus-feqa}
Esin Durmus, He~He, and Mona Diab. 2020.
\newblock \href {https://doi.org/10.18653/v1/2020.acl-main.454} {{FEQA}: A question answering evaluation framework for faithfulness assessment in abstractive summarization}.
\newblock In \emph{Proceedings of the 58th Annual Meeting of the Association for Computational Linguistics}, pages 5055--5070, Online. Association for Computational Linguistics.

\bibitem[{Fabbri et~al.(2022)Fabbri, Wu, Liu, and Xiong}]{fabbri-qafacteval}
Alexander Fabbri, Chien-Sheng Wu, Wenhao Liu, and Caiming Xiong. 2022.
\newblock \href {https://doi.org/10.18653/v1/2022.naacl-main.187} {{QAF}act{E}val: Improved {QA}-based factual consistency evaluation for summarization}.
\newblock In \emph{Proceedings of the 2022 Conference of the North American Chapter of the Association for Computational Linguistics: Human Language Technologies}, pages 2587--2601, Seattle, United States. Association for Computational Linguistics.

\bibitem[{Feng et~al.(2024)Feng, Sorensen, Liu, Fisher, Park, Choi, and Tsvetkov}]{feng2024modular}
Shangbin Feng, Taylor Sorensen, Yuhan Liu, Jillian Fisher, Chan~Young Park, Yejin Choi, and Yulia Tsvetkov. 2024.
\newblock \href {https://doi.org/10.18653/v1/2024.emnlp-main.240} {Modular pluralism: Pluralistic alignment via multi-{LLM} collaboration}.
\newblock In \emph{Proceedings of the 2024 Conference on Empirical Methods in Natural Language Processing}, pages 4151--4171, Miami, Florida, USA. Association for Computational Linguistics.

\bibitem[{Gooding and Mansoor(2023)}]{gooding2023impact}
Sian Gooding and Hassan Mansoor. 2023.
\newblock \href {https://arxiv.org/abs/2311.04919} {The impact of preference agreement in reinforcement learning from human feedback: A case study in summarization}.
\newblock \emph{Preprint}, arXiv:2311.04919.

\bibitem[{Goyal et~al.(2023)Goyal, Li, and Durrett}]{goyal2023news}
Tanya Goyal, Junyi~Jessy Li, and Greg Durrett. 2023.
\newblock \href {https://arxiv.org/abs/2209.12356} {News summarization and evaluation in the era of gpt-3}.
\newblock \emph{Preprint}, arXiv:2209.12356.

\bibitem[{Grusky et~al.(2018)Grusky, Naaman, and Artzi}]{grusky2018newsroom}
Max Grusky, Mor Naaman, and Yoav Artzi. 2018.
\newblock \href {https://doi.org/10.18653/v1/N18-1065} {{N}ewsroom: A dataset of 1.3 million summaries with diverse extractive strategies}.
\newblock In \emph{Proceedings of the 2018 Conference of the North {A}merican Chapter of the Association for Computational Linguistics: Human Language Technologies, Volume 1 (Long Papers)}, pages 708--719, New Orleans, Louisiana. Association for Computational Linguistics.

\bibitem[{Horvitz et~al.(2024)Horvitz, Patel, Singh, Callison-Burch, McKeown, and Yu}]{horvitz2024tinystyler}
Zachary Horvitz, Ajay Patel, Kanishk Singh, Chris Callison-Burch, Kathleen McKeown, and Zhou Yu. 2024.
\newblock \href {https://doi.org/10.18653/v1/2024.findings-emnlp.781} {{T}iny{S}tyler: Efficient few-shot text style transfer with authorship embeddings}.
\newblock In \emph{Findings of the Association for Computational Linguistics: EMNLP 2024}, pages 13376--13390, Miami, Florida, USA. Association for Computational Linguistics.

\bibitem[{Hsieh et~al.(2024)Hsieh, Chuang, Li, Wang, Le, Kumar, Glass, Ratner, Lee, Krishna, and Pfister}]{hsieh2024found}
Cheng-Yu Hsieh, Yung-Sung Chuang, Chun-Liang Li, Zifeng Wang, Long Le, Abhishek Kumar, James Glass, Alexander Ratner, Chen-Yu Lee, Ranjay Krishna, and Tomas Pfister. 2024.
\newblock \href {https://doi.org/10.18653/v1/2024.findings-acl.890} {Found in the middle: Calibrating positional attention bias improves long context utilization}.
\newblock In \emph{Findings of the Association for Computational Linguistics: ACL 2024}, pages 14982--14995, Bangkok, Thailand. Association for Computational Linguistics.

\bibitem[{Hu et~al.(2024)Hu, Song, Zhang, Xiao, Wang, Chen, Yuan, Lian, Ding, and Xiong}]{hu2024explaining}
Zhengyu Hu, Linxin Song, Jieyu Zhang, Zheyuan Xiao, Tianfu Wang, Zhengyu Chen, Nicholas~Jing Yuan, Jianxun Lian, Kaize Ding, and Hui Xiong. 2024.
\newblock \href {https://arxiv.org/abs/2407.01085} {Explaining length bias in llm-based preference evaluations}.
\newblock \emph{Preprint}, arXiv:2407.01085.

\bibitem[{Huang et~al.(2023)Huang, Gu, Hou, Wu, Wang, Yu, and Han}]{huang2023large}
Jiaxin Huang, Shixiang Gu, Le~Hou, Yuexin Wu, Xuezhi Wang, Hongkun Yu, and Jiawei Han. 2023.
\newblock \href {https://doi.org/10.18653/v1/2023.emnlp-main.67} {Large language models can self-improve}.
\newblock In \emph{Proceedings of the 2023 Conference on Empirical Methods in Natural Language Processing}, pages 1051--1068, Singapore. Association for Computational Linguistics.

\bibitem[{Huang et~al.(2024)Huang, Noukhovitch, Hosseini, Rasul, Wang, and Tunstall}]{huang2024the}
Shengyi Huang, Michael Noukhovitch, Arian Hosseini, Kashif Rasul, Weixun Wang, and Lewis Tunstall. 2024.
\newblock \href {https://openreview.net/forum?id=kHO2ZTa8e3} {The n+ implementation details of {RLHF} with {PPO}: A case study on {TL};{DR} summarization}.
\newblock In \emph{First Conference on Language Modeling}.

\bibitem[{Iso et~al.(2022)Iso, Wang, Angelidis, and Suhara}]{iso2022comparative}
Hayate Iso, Xiaolan Wang, Stefanos Angelidis, and Yoshihiko Suhara. 2022.
\newblock \href {https://doi.org/10.18653/v1/2022.findings-acl.261} {Comparative opinion summarization via collaborative decoding}.
\newblock In \emph{Findings of the Association for Computational Linguistics: ACL 2022}, pages 3307--3324, Dublin, Ireland. Association for Computational Linguistics.

\bibitem[{Jacob et~al.(2024)Jacob, Shen, Farina, and Andreas}]{jacob2024the}
Athul~Paul Jacob, Yikang Shen, Gabriele Farina, and Jacob Andreas. 2024.
\newblock \href {https://openreview.net/forum?id=n9xeGcI4Yg} {The consensus game: Language model generation via equilibrium search}.
\newblock In \emph{The Twelfth International Conference on Learning Representations}.

\bibitem[{Jain et~al.(2023)Jain, Keshava, Mysore~Sathyendra, Fernandes, Liu, Neubig, and Zhou}]{jain2023multi}
Sameer Jain, Vaishakh Keshava, Swarnashree Mysore~Sathyendra, Patrick Fernandes, Pengfei Liu, Graham Neubig, and Chunting Zhou. 2023.
\newblock \href {https://doi.org/10.18653/v1/2023.findings-acl.537} {Multi-dimensional evaluation of text summarization with in-context learning}.
\newblock In \emph{Findings of the Association for Computational Linguistics: ACL 2023}, pages 8487--8495, Toronto, Canada. Association for Computational Linguistics.

\bibitem[{Jung et~al.(2019)Jung, Kang, Mentch, and Hovy}]{jung2019earlier}
Taehee Jung, Dongyeop Kang, Lucas Mentch, and Eduard Hovy. 2019.
\newblock \href {https://arxiv.org/abs/1908.11723} {Earlier isn't always better: Sub-aspect analysis on corpus and system biases in summarization}.
\newblock \emph{Preprint}, arXiv:1908.11723.

\bibitem[{Kryscinski et~al.(2020)Kryscinski, McCann, Xiong, and Socher}]{kryscinski-factcc}
Wojciech Kryscinski, Bryan McCann, Caiming Xiong, and Richard Socher. 2020.
\newblock \href {https://doi.org/10.18653/v1/2020.emnlp-main.750} {Evaluating the factual consistency of abstractive text summarization}.
\newblock In \emph{Proceedings of the 2020 Conference on Empirical Methods in Natural Language Processing (EMNLP)}, pages 9332--9346, Online. Association for Computational Linguistics.

\bibitem[{Laban et~al.(2022)Laban, Schnabel, Bennett, and Hearst}]{laban2022summac}
Philippe Laban, Tobias Schnabel, Paul~N. Bennett, and Marti~A. Hearst. 2022.
\newblock \href {https://doi.org/10.1162/tacl_a_00453} {{S}umma{C}: Re-visiting {NLI}-based models for inconsistency detection in summarization}.
\newblock \emph{Transactions of the Association for Computational Linguistics}, 10:163--177.

\bibitem[{Ladhak et~al.(2022)Ladhak, Durmus, He, Cardie, and McKeown}]{ladhak-faithful}
Faisal Ladhak, Esin Durmus, He~He, Claire Cardie, and Kathleen McKeown. 2022.
\newblock \href {https://doi.org/10.18653/v1/2022.acl-long.100} {Faithful or extractive? on mitigating the faithfulness-abstractiveness trade-off in abstractive summarization}.
\newblock In \emph{Proceedings of the 60th Annual Meeting of the Association for Computational Linguistics (Volume 1: Long Papers)}, pages 1410--1421, Dublin, Ireland. Association for Computational Linguistics.

\bibitem[{Lee et~al.(2024)Lee, Phatale, Mansoor, Lu, Mesnard, Ferret, Bishop, Hall, Carbune, and Rastogi}]{lee2024rlaif}
Harrison Lee, Samrat Phatale, Hassan Mansoor, Kellie~Ren Lu, Thomas Mesnard, Johan Ferret, Colton Bishop, Ethan Hall, Victor Carbune, and Abhinav Rastogi. 2024.
\newblock \href {https://openreview.net/forum?id=AAxIs3D2ZZ} {{RLAIF}: Scaling reinforcement learning from human feedback with {AI} feedback}.

\bibitem[{Lee et~al.(2022{\natexlab{a}})Lee, Bang, Yu, Madotto, and Fung}]{lee2022neus}
Nayeon Lee, Yejin Bang, Tiezheng Yu, Andrea Madotto, and Pascale Fung. 2022{\natexlab{a}}.
\newblock \href {https://doi.org/10.18653/v1/2022.naacl-main.228} {{N}eu{S}: Neutral multi-news summarization for mitigating framing bias}.
\newblock In \emph{Proceedings of the 2022 Conference of the North American Chapter of the Association for Computational Linguistics: Human Language Technologies}, pages 3131--3148, Seattle, United States. Association for Computational Linguistics.

\bibitem[{Lee et~al.(2022{\natexlab{b}})Lee, Bang, Yu, Madotto, and Fung}]{lee-neus}
Nayeon Lee, Yejin Bang, Tiezheng Yu, Andrea Madotto, and Pascale Fung. 2022{\natexlab{b}}.
\newblock \href {https://doi.org/10.18653/v1/2022.naacl-main.228} {{N}eu{S}: Neutral multi-news summarization for mitigating framing bias}.
\newblock In \emph{Proceedings of the 2022 Conference of the North American Chapter of the Association for Computational Linguistics: Human Language Technologies}, pages 3131--3148, Seattle, United States. Association for Computational Linguistics.

\bibitem[{Lei et~al.(2024)Lei, Song, Cho, Wang, Huang, and Yu}]{lei2024polarity}
Yuanyuan Lei, Kaiqiang Song, Sangwoo Cho, Xiaoyang Wang, Ruihong Huang, and Dong Yu. 2024.
\newblock \href {https://doi.org/10.18653/v1/2024.naacl-long.291} {Polarity calibration for opinion summarization}.
\newblock In \emph{Proceedings of the 2024 Conference of the North American Chapter of the Association for Computational Linguistics: Human Language Technologies (Volume 1: Long Papers)}, pages 5211--5224, Mexico City, Mexico. Association for Computational Linguistics.

\bibitem[{Lin(2004)}]{lin2004rouge}
Chin-Yew Lin. 2004.
\newblock \href {https://aclanthology.org/W04-1013/} {{ROUGE}: A package for automatic evaluation of summaries}.
\newblock In \emph{Text Summarization Branches Out}, pages 74--81, Barcelona, Spain. Association for Computational Linguistics.

\bibitem[{Liu et~al.(2024{\natexlab{a}})Liu, Lin, Hewitt, Paranjape, Bevilacqua, Petroni, and Liang}]{liu2024lost}
Nelson~F. Liu, Kevin Lin, John Hewitt, Ashwin Paranjape, Michele Bevilacqua, Fabio Petroni, and Percy Liang. 2024{\natexlab{a}}.
\newblock \href {https://doi.org/10.1162/tacl_a_00638} {Lost in the middle: How language models use long contexts}.
\newblock \emph{Transactions of the Association for Computational Linguistics}, 12:157--173.

\bibitem[{Liu et~al.(2023)Liu, Iter, Xu, Wang, Xu, and Zhu}]{liu2023g}
Yang Liu, Dan Iter, Yichong Xu, Shuohang Wang, Ruochen Xu, and Chenguang Zhu. 2023.
\newblock \href {https://doi.org/10.18653/v1/2023.emnlp-main.153} {{G}-eval: {NLG} evaluation using gpt-4 with better human alignment}.
\newblock In \emph{Proceedings of the 2023 Conference on Empirical Methods in Natural Language Processing}, pages 2511--2522, Singapore. Association for Computational Linguistics.

\bibitem[{Liu et~al.(2024{\natexlab{b}})Liu, Feng, Han, Balachandran, Park, Kumar, and Tsvetkov}]{liu2024p3sum}
Yuhan Liu, Shangbin Feng, Xiaochuang Han, Vidhisha Balachandran, Chan~Young Park, Sachin Kumar, and Yulia Tsvetkov. 2024{\natexlab{b}}.
\newblock \href {https://doi.org/10.18653/v1/2024.naacl-long.119} {{P}$^3${S}um: Preserving author`s perspective in news summarization with diffusion language models}.
\newblock In \emph{Proceedings of the 2024 Conference of the North American Chapter of the Association for Computational Linguistics: Human Language Technologies (Volume 1: Long Papers)}, pages 2154--2173, Mexico City, Mexico. Association for Computational Linguistics.

\bibitem[{Madaan et~al.(2023)Madaan, Tandon, Gupta, Hallinan, Gao, Wiegreffe, Alon, Dziri, Prabhumoye, Yang, Gupta, Majumder, Hermann, Welleck, Yazdanbakhsh, and Clark}]{madaan2023selfrefine}
Aman Madaan, Niket Tandon, Prakhar Gupta, Skyler Hallinan, Luyu Gao, Sarah Wiegreffe, Uri Alon, Nouha Dziri, Shrimai Prabhumoye, Yiming Yang, Shashank Gupta, Bodhisattwa~Prasad Majumder, Katherine Hermann, Sean Welleck, Amir Yazdanbakhsh, and Peter Clark. 2023.
\newblock \href {https://openreview.net/forum?id=S37hOerQLB} {Self-refine: Iterative refinement with self-feedback}.
\newblock In \emph{Thirty-seventh Conference on Neural Information Processing Systems}.

\bibitem[{Maynez et~al.(2020)Maynez, Narayan, Bohnet, and McDonald}]{maynez2020faithfulness}
Joshua Maynez, Shashi Narayan, Bernd Bohnet, and Ryan McDonald. 2020.
\newblock \href {https://doi.org/10.18653/v1/2020.acl-main.173} {On faithfulness and factuality in abstractive summarization}.
\newblock In \emph{Proceedings of the 58th Annual Meeting of the Association for Computational Linguistics}, pages 1906--1919, Online. Association for Computational Linguistics.

\bibitem[{Nakano et~al.(2022)Nakano, Hilton, Balaji, Wu, Ouyang, Kim, Hesse, Jain, Kosaraju, Saunders, Jiang, Cobbe, Eloundou, Krueger, Button, Knight, Chess, and Schulman}]{nakano2022webgpt}
Reiichiro Nakano, Jacob Hilton, Suchir Balaji, Jeff Wu, Long Ouyang, Christina Kim, Christopher Hesse, Shantanu Jain, Vineet Kosaraju, William Saunders, Xu~Jiang, Karl Cobbe, Tyna Eloundou, Gretchen Krueger, Kevin Button, Matthew Knight, Benjamin Chess, and John Schulman. 2022.
\newblock \href {https://arxiv.org/abs/2112.09332} {Webgpt: Browser-assisted question-answering with human feedback}.
\newblock \emph{Preprint}, arXiv:2112.09332.

\bibitem[{Nan et~al.(2021)Nan, Nogueira~dos Santos, Zhu, Ng, McKeown, Nallapati, Zhang, Wang, Arnold, and Xiang}]{nan-contrastive}
Feng Nan, Cicero Nogueira~dos Santos, Henghui Zhu, Patrick Ng, Kathleen McKeown, Ramesh Nallapati, Dejiao Zhang, Zhiguo Wang, Andrew~O. Arnold, and Bing Xiang. 2021.
\newblock \href {https://doi.org/10.18653/v1/2021.acl-long.536} {Improving factual consistency of abstractive summarization via question answering}.
\newblock In \emph{Proceedings of the 59th Annual Meeting of the Association for Computational Linguistics and the 11th International Joint Conference on Natural Language Processing (Volume 1: Long Papers)}, pages 6881--6894, Online. Association for Computational Linguistics.

\bibitem[{Ouyang et~al.(2022)Ouyang, Wu, Jiang, Almeida, Wainwright, Mishkin, Zhang, Agarwal, Slama, Ray, Schulman, Hilton, Kelton, Miller, Simens, Askell, Welinder, Christiano, Leike, and Lowe}]{ouyang2022training}
Long Ouyang, Jeffrey Wu, Xu~Jiang, Diogo Almeida, Carroll Wainwright, Pamela Mishkin, Chong Zhang, Sandhini Agarwal, Katarina Slama, Alex Ray, John Schulman, Jacob Hilton, Fraser Kelton, Luke Miller, Maddie Simens, Amanda Askell, Peter Welinder, Paul~F Christiano, Jan Leike, and Ryan Lowe. 2022.
\newblock \href {https://proceedings.neurips.cc/paper_files/paper/2022/file/b1efde53be364a73914f58805a001731-Paper-Conference.pdf} {Training language models to follow instructions with human feedback}.
\newblock In \emph{Advances in Neural Information Processing Systems}, volume~35, pages 27730--27744. Curran Associates, Inc.

\bibitem[{Papineni et~al.(2002)Papineni, Roukos, Ward, and Zhu}]{papineni-bleu}
Kishore Papineni, Salim Roukos, Todd Ward, and Wei-Jing Zhu. 2002.
\newblock \href {https://doi.org/10.3115/1073083.1073135} {{B}leu: a method for automatic evaluation of machine translation}.
\newblock In \emph{Proceedings of the 40th Annual Meeting of the Association for Computational Linguistics}, pages 311--318, Philadelphia, Pennsylvania, USA. Association for Computational Linguistics.

\bibitem[{Parcalabescu and Frank(2024)}]{parcalabescu2024measuring}
Letitia Parcalabescu and Anette Frank. 2024.
\newblock \href {https://doi.org/10.18653/v1/2024.acl-long.329} {On measuring faithfulness or self-consistency of natural language explanations}.
\newblock In \emph{Proceedings of the 62nd Annual Meeting of the Association for Computational Linguistics (Volume 1: Long Papers)}, pages 6048--6089, Bangkok, Thailand. Association for Computational Linguistics.

\bibitem[{Popovi{\'c}(2015)}]{popovic-2015-chrf}
Maja Popovi{\'c}. 2015.
\newblock \href {https://doi.org/10.18653/v1/W15-3049} {chr{F}: character n-gram {F}-score for automatic {MT} evaluation}.
\newblock In \emph{Proceedings of the Tenth Workshop on Statistical Machine Translation}, pages 392--395, Lisbon, Portugal. Association for Computational Linguistics.

\bibitem[{Press et~al.(2023)Press, Zhang, Min, Schmidt, Smith, and Lewis}]{press2023measuring}
Ofir Press, Muru Zhang, Sewon Min, Ludwig Schmidt, Noah Smith, and Mike Lewis. 2023.
\newblock \href {https://doi.org/10.18653/v1/2023.findings-emnlp.378} {Measuring and narrowing the compositionality gap in language models}.
\newblock In \emph{Findings of the Association for Computational Linguistics: EMNLP 2023}, pages 5687--5711, Singapore. Association for Computational Linguistics.

\bibitem[{Rafailov et~al.(2023)Rafailov, Sharma, Mitchell, Manning, Ermon, and Finn}]{rafailov2023direct}
Rafael Rafailov, Archit Sharma, Eric Mitchell, Christopher~D Manning, Stefano Ermon, and Chelsea Finn. 2023.
\newblock \href {https://openreview.net/forum?id=HPuSIXJaa9} {Direct preference optimization: Your language model is secretly a reward model}.
\newblock In \emph{Thirty-seventh Conference on Neural Information Processing Systems}.

\bibitem[{Ratner et~al.(2023)Ratner, Levine, Belinkov, Ram, Magar, Abend, Karpas, Shashua, Leyton-Brown, and Shoham}]{ratner2023parallel}
Nir Ratner, Yoav Levine, Yonatan Belinkov, Ori Ram, Inbal Magar, Omri Abend, Ehud Karpas, Amnon Shashua, Kevin Leyton-Brown, and Yoav Shoham. 2023.
\newblock \href {https://doi.org/10.18653/v1/2023.acl-long.352} {Parallel context windows for large language models}.
\newblock In \emph{Proceedings of the 61st Annual Meeting of the Association for Computational Linguistics (Volume 1: Long Papers)}, pages 6383--6402, Toronto, Canada. Association for Computational Linguistics.

\bibitem[{Roit et~al.(2023)Roit, Ferret, Shani, Aharoni, Cideron, Dadashi, Geist, Girgin, Hussenot, Keller, Momchev, Ramos~Garea, Stanczyk, Vieillard, Bachem, Elidan, Hassidim, Pietquin, and Szpektor}]{roit-rlef}
Paul Roit, Johan Ferret, Lior Shani, Roee Aharoni, Geoffrey Cideron, Robert Dadashi, Matthieu Geist, Sertan Girgin, Leonard Hussenot, Orgad Keller, Nikola Momchev, Sabela Ramos~Garea, Piotr Stanczyk, Nino Vieillard, Olivier Bachem, Gal Elidan, Avinatan Hassidim, Olivier Pietquin, and Idan Szpektor. 2023.
\newblock \href {https://doi.org/10.18653/v1/2023.acl-long.344} {Factually consistent summarization via reinforcement learning with textual entailment feedback}.
\newblock In \emph{Proceedings of the 61st Annual Meeting of the Association for Computational Linguistics (Volume 1: Long Papers)}, pages 6252--6272, Toronto, Canada. Association for Computational Linguistics.

\bibitem[{Saha et~al.(2024)Saha, Levy, Celikyilmaz, Bansal, Weston, and Li}]{saha2024branch}
Swarnadeep Saha, Omer Levy, Asli Celikyilmaz, Mohit Bansal, Jason Weston, and Xian Li. 2024.
\newblock \href {https://doi.org/10.18653/v1/2024.naacl-long.462} {Branch-solve-merge improves large language model evaluation and generation}.
\newblock In \emph{Proceedings of the 2024 Conference of the North American Chapter of the Association for Computational Linguistics: Human Language Technologies (Volume 1: Long Papers)}, pages 8352--8370, Mexico City, Mexico. Association for Computational Linguistics.

\bibitem[{See et~al.(2017)See, Liu, and Manning}]{see2017get}
Abigail See, Peter~J. Liu, and Christopher~D. Manning. 2017.
\newblock \href {https://doi.org/10.18653/v1/P17-1099} {Get to the point: Summarization with pointer-generator networks}.
\newblock In \emph{Proceedings of the 55th Annual Meeting of the Association for Computational Linguistics (Volume 1: Long Papers)}, pages 1073--1083, Vancouver, Canada. Association for Computational Linguistics.

\bibitem[{Sellam et~al.(2020)Sellam, Das, and Parikh}]{sellam2020bleurt}
Thibault Sellam, Dipanjan Das, and Ankur Parikh. 2020.
\newblock \href {https://doi.org/10.18653/v1/2020.acl-main.704} {{BLEURT}: Learning robust metrics for text generation}.
\newblock In \emph{Proceedings of the 58th Annual Meeting of the Association for Computational Linguistics}, pages 7881--7892, Online. Association for Computational Linguistics.

\bibitem[{Siegel et~al.(2024)Siegel, Camburu, Heess, and Perez-Ortiz}]{siegel2024probabilities}
Noah Siegel, Oana-Maria Camburu, Nicolas Heess, and Maria Perez-Ortiz. 2024.
\newblock \href {https://doi.org/10.18653/v1/2024.acl-short.49} {The probabilities also matter: A more faithful metric for faithfulness of free-text explanations in large language models}.
\newblock In \emph{Proceedings of the 62nd Annual Meeting of the Association for Computational Linguistics (Volume 2: Short Papers)}, pages 530--546, Bangkok, Thailand. Association for Computational Linguistics.

\bibitem[{Song et~al.(2024)Song, Su, Shalyminov, Cai, and Mansour}]{song2024finesure}
Hwanjun Song, Hang Su, Igor Shalyminov, Jason Cai, and Saab Mansour. 2024.
\newblock \href {https://doi.org/10.18653/v1/2024.acl-long.51} {{F}ine{S}ur{E}: Fine-grained summarization evaluation using {LLM}s}.
\newblock In \emph{Proceedings of the 62nd Annual Meeting of the Association for Computational Linguistics (Volume 1: Long Papers)}, pages 906--922, Bangkok, Thailand. Association for Computational Linguistics.

\bibitem[{Stiennon et~al.(2020)Stiennon, Ouyang, Wu, Ziegler, Lowe, Voss, Radford, Amodei, and Christiano}]{stiennon2020learning}
Nisan Stiennon, Long Ouyang, Jeffrey Wu, Daniel Ziegler, Ryan Lowe, Chelsea Voss, Alec Radford, Dario Amodei, and Paul~F Christiano. 2020.
\newblock \href {https://proceedings.neurips.cc/paper_files/paper/2020/file/1f89885d556929e98d3ef9b86448f951-Paper.pdf} {Learning to summarize with human feedback}.
\newblock In \emph{Advances in Neural Information Processing Systems}, volume~33, pages 3008--3021. Curran Associates, Inc.

\bibitem[{Suzgun et~al.(2022)Suzgun, Melas-Kyriazi, and Jurafsky}]{suzgun2022prompt}
Mirac Suzgun, Luke Melas-Kyriazi, and Dan Jurafsky. 2022.
\newblock \href {https://doi.org/10.18653/v1/2022.emnlp-main.141} {Prompt-and-rerank: A method for zero-shot and few-shot arbitrary textual style transfer with small language models}.
\newblock In \emph{Proceedings of the 2022 Conference on Empirical Methods in Natural Language Processing}, pages 2195--2222, Abu Dhabi, United Arab Emirates. Association for Computational Linguistics.

\bibitem[{Tang et~al.(2024)Tang, Laban, and Durrett}]{tang2024minicheck}
Liyan Tang, Philippe Laban, and Greg Durrett. 2024.
\newblock \href {https://doi.org/10.18653/v1/2024.emnlp-main.499} {{M}ini{C}heck: Efficient fact-checking of {LLM}s on grounding documents}.
\newblock In \emph{Proceedings of the 2024 Conference on Empirical Methods in Natural Language Processing}, pages 8818--8847, Miami, Florida, USA. Association for Computational Linguistics.

\bibitem[{Vijayakumar et~al.(2018)Vijayakumar, Cogswell, Selvaraju, Sun, Lee, Crandall, and Batra}]{vijayakumar2018diverse}
Ashwin Vijayakumar, Michael Cogswell, Ramprasaath Selvaraju, Qing Sun, Stefan Lee, David Crandall, and Dhruv Batra. 2018.
\newblock \href {https://doi.org/10.1609/aaai.v32i1.12340} {Diverse beam search for improved description of complex scenes}.
\newblock \emph{Proceedings of the AAAI Conference on Artificial Intelligence}, 32(1).

\bibitem[{Wang et~al.(2020)Wang, Cho, and Lewis}]{wang-qags}
Alex Wang, Kyunghyun Cho, and Mike Lewis. 2020.
\newblock \href {https://doi.org/10.18653/v1/2020.acl-main.450} {Asking and answering questions to evaluate the factual consistency of summaries}.
\newblock In \emph{Proceedings of the 58th Annual Meeting of the Association for Computational Linguistics}, pages 5008--5020, Online. Association for Computational Linguistics.

\bibitem[{Wang et~al.(2023{\natexlab{a}})Wang, Wei, Schuurmans, Le, Chi, Narang, Chowdhery, and Zhou}]{wang2023selfconsistency}
Xuezhi Wang, Jason Wei, Dale Schuurmans, Quoc~V Le, Ed~H. Chi, Sharan Narang, Aakanksha Chowdhery, and Denny Zhou. 2023{\natexlab{a}}.
\newblock \href {https://openreview.net/forum?id=1PL1NIMMrw} {Self-consistency improves chain of thought reasoning in language models}.
\newblock In \emph{The Eleventh International Conference on Learning Representations}.

\bibitem[{Wang et~al.(2023{\natexlab{b}})Wang, Zhang, and Wang}]{wang2023element}
Yiming Wang, Zhuosheng Zhang, and Rui Wang. 2023{\natexlab{b}}.
\newblock \href {https://doi.org/10.18653/v1/2023.acl-long.482} {Element-aware summarization with large language models: Expert-aligned evaluation and chain-of-thought method}.
\newblock In \emph{Proceedings of the 61st Annual Meeting of the Association for Computational Linguistics (Volume 1: Long Papers)}, pages 8640--8665, Toronto, Canada. Association for Computational Linguistics.

\bibitem[{Wang et~al.(2024)Wang, Zhang, Li, Huang, Han, Ji, Kakade, Peng, and Ji}]{wang2024eliminating}
Ziqi Wang, Hanlin Zhang, Xiner Li, Kuan-Hao Huang, Chi Han, Shuiwang Ji, Sham~M. Kakade, Hao Peng, and Heng Ji. 2024.
\newblock \href {https://arxiv.org/abs/2407.01100} {Eliminating position bias of language models: A mechanistic approach}.
\newblock \emph{Preprint}, arXiv:2407.01100.

\bibitem[{Wei et~al.(2022)Wei, Wang, Schuurmans, Bosma, brian ichter, Xia, Chi, Le, and Zhou}]{wei2022chain}
Jason Wei, Xuezhi Wang, Dale Schuurmans, Maarten Bosma, brian ichter, Fei Xia, Ed~H. Chi, Quoc~V Le, and Denny Zhou. 2022.
\newblock \href {https://openreview.net/forum?id=_VjQlMeSB_J} {Chain of thought prompting elicits reasoning in large language models}.
\newblock In \emph{Advances in Neural Information Processing Systems}.

\bibitem[{Weng et~al.(2023)Weng, Zhu, Xia, Li, He, Liu, Sun, Liu, and Zhao}]{weng2023large}
Yixuan Weng, Minjun Zhu, Fei Xia, Bin Li, Shizhu He, Shengping Liu, Bin Sun, Kang Liu, and Jun Zhao. 2023.
\newblock \href {https://doi.org/10.18653/v1/2023.findings-emnlp.167} {Large language models are better reasoners with self-verification}.
\newblock In \emph{Findings of the Association for Computational Linguistics: EMNLP 2023}, pages 2550--2575, Singapore. Association for Computational Linguistics.

\bibitem[{Wu et~al.(2024)Wu, Iso, Pezeshkpour, Bhutani, and Hruschka}]{wu2024less}
Yunshu Wu, Hayate Iso, Pouya Pezeshkpour, Nikita Bhutani, and Estevam Hruschka. 2024.
\newblock \href {https://aclanthology.org/2024.eacl-short.29/} {Less is more for long document summary evaluation by {LLM}s}.
\newblock In \emph{Proceedings of the 18th Conference of the European Chapter of the Association for Computational Linguistics (Volume 2: Short Papers)}, pages 330--343, St. Julian{'}s, Malta. Association for Computational Linguistics.

\bibitem[{Xu et~al.(2024)Xu, Deutsch, Finkelstein, Juraska, Zhang, Liu, Wang, Li, and Freitag}]{xu2024llmrefine}
Wenda Xu, Daniel Deutsch, Mara Finkelstein, Juraj Juraska, Biao Zhang, Zhongtao Liu, William~Yang Wang, Lei Li, and Markus Freitag. 2024.
\newblock \href {https://doi.org/10.18653/v1/2024.findings-naacl.92} {{LLMR}efine: Pinpointing and refining large language models via fine-grained actionable feedback}.
\newblock In \emph{Findings of the Association for Computational Linguistics: NAACL 2024}, pages 1429--1445, Mexico City, Mexico. Association for Computational Linguistics.

\bibitem[{Zha et~al.(2023)Zha, Yang, Li, and Hu}]{zha2023alignscore}
Yuheng Zha, Yichi Yang, Ruichen Li, and Zhiting Hu. 2023.
\newblock \href {https://doi.org/10.18653/v1/2023.acl-long.634} {{A}lign{S}core: Evaluating factual consistency with a unified alignment function}.
\newblock In \emph{Proceedings of the 61st Annual Meeting of the Association for Computational Linguistics (Volume 1: Long Papers)}, pages 11328--11348, Toronto, Canada. Association for Computational Linguistics.

\bibitem[{Zhang and Bansal(2021)}]{zhang2021finding}
Shiyue Zhang and Mohit Bansal. 2021.
\newblock \href {https://doi.org/10.18653/v1/2021.emnlp-main.531} {Finding a balanced degree of automation for summary evaluation}.
\newblock In \emph{Proceedings of the 2021 Conference on Empirical Methods in Natural Language Processing}, pages 6617--6632, Online and Punta Cana, Dominican Republic. Association for Computational Linguistics.

\bibitem[{Zhang et~al.(2020)Zhang, Kishore, Wu, Weinberger, and Artzi}]{zhang2020bertscore}
Tianyi Zhang, Varsha Kishore, Felix Wu, Kilian~Q. Weinberger, and Yoav Artzi. 2020.
\newblock \href {https://openreview.net/forum?id=SkeHuCVFDr} {Bertscore: Evaluating text generation with bert}.
\newblock In \emph{International Conference on Learning Representations}.

\bibitem[{Zhang et~al.(2024{\natexlab{a}})Zhang, Ladhak, Durmus, Liang, McKeown, and Hashimoto}]{zhang2024benchmarking}
Tianyi Zhang, Faisal Ladhak, Esin Durmus, Percy Liang, Kathleen McKeown, and Tatsunori~B. Hashimoto. 2024{\natexlab{a}}.
\newblock \href {https://doi.org/10.1162/tacl_a_00632} {Benchmarking large language models for news summarization}.
\newblock \emph{Transactions of the Association for Computational Linguistics}, 12:39--57.

\bibitem[{Zhang et~al.(2024{\natexlab{b}})Zhang, Shen, Wu, Peng, Wang, Zhuang, and Lu}]{zhang2024selfcontrast}
Wenqi Zhang, Yongliang Shen, Linjuan Wu, Qiuying Peng, Jun Wang, Yueting Zhuang, and Weiming Lu. 2024{\natexlab{b}}.
\newblock \href {https://doi.org/10.18653/v1/2024.acl-long.197} {Self-contrast: Better reflection through inconsistent solving perspectives}.
\newblock In \emph{Proceedings of the 62nd Annual Meeting of the Association for Computational Linguistics (Volume 1: Long Papers)}, pages 3602--3622, Bangkok, Thailand. Association for Computational Linguistics.

\bibitem[{Zhang et~al.(2024{\natexlab{c}})Zhang, Zhang, Liu, Fabbri, Liu, Kamoi, Lu, Xiong, Zhao, Radev, McKeown, and Zhang}]{zhang2024fair}
Yusen Zhang, Nan Zhang, Yixin Liu, Alexander Fabbri, Junru Liu, Ryo Kamoi, Xiaoxin Lu, Caiming Xiong, Jieyu Zhao, Dragomir Radev, Kathleen McKeown, and Rui Zhang. 2024{\natexlab{c}}.
\newblock \href {https://doi.org/10.18653/v1/2024.naacl-long.187} {Fair abstractive summarization of diverse perspectives}.
\newblock In \emph{Proceedings of the 2024 Conference of the North American Chapter of the Association for Computational Linguistics: Human Language Technologies (Volume 1: Long Papers)}, pages 3404--3426, Mexico City, Mexico. Association for Computational Linguistics.

\bibitem[{Zheng et~al.(2023)Zheng, Chiang, Sheng, Zhuang, Wu, Zhuang, Lin, Li, Li, Xing, Zhang, Gonzalez, and Stoica}]{zheng2023judging}
Lianmin Zheng, Wei-Lin Chiang, Ying Sheng, Siyuan Zhuang, Zhanghao Wu, Yonghao Zhuang, Zi~Lin, Zhuohan Li, Dacheng Li, Eric Xing, Hao Zhang, Joseph~E. Gonzalez, and Ion Stoica. 2023.
\newblock \href {https://openreview.net/forum?id=uccHPGDlao} {Judging {LLM}-as-a-judge with {MT}-bench and chatbot arena}.
\newblock In \emph{Thirty-seventh Conference on Neural Information Processing Systems Datasets and Benchmarks Track}.

\bibitem[{Zhong et~al.(2022)Zhong, Liu, Yin, Mao, Jiao, Liu, Zhu, Ji, and Han}]{zhong2022towards}
Ming Zhong, Yang Liu, Da~Yin, Yuning Mao, Yizhu Jiao, Pengfei Liu, Chenguang Zhu, Heng Ji, and Jiawei Han. 2022.
\newblock \href {https://doi.org/10.18653/v1/2022.emnlp-main.131} {Towards a unified multi-dimensional evaluator for text generation}.
\newblock In \emph{Proceedings of the 2022 Conference on Empirical Methods in Natural Language Processing}, pages 2023--2038, Abu Dhabi, United Arab Emirates. Association for Computational Linguistics.

\end{thebibliography}

\appendix
\section{Supplementary Details}
\label{app:supplementary_details}

\begin{figure}[t]
    \centering
    \begin{lstlisting}[basicstyle=\ttfamily\scriptsize, 
                       frame=single, 
                       breaklines=true,
                       breakatwhitespace=false,
                       columns=fullflexible,
                       keepspaces=true,
                       xleftmargin=0pt,
                       xrightmargin=0pt,
                       breakindent=0ex]
Given texts from both Left-leaning and Right-leaning perspectives, summarize only the Left-leaning perspective in one sentence, starting with 'The Left '. ONLY RETURN THE SUMMARY AND NOTHING ELSE.

Left:
(left-perspective article)

Right:
(right-perspective article)\end{lstlisting}
    \vspace{-5mm}
    \caption{Prompt instruction for zero-shot inference when generating summaries from the left-leaning perspective.}
    \label{fig:zero_shot_prompt}
\end{figure}

\subsection{Experimental Setup}
\label{app:experimental_setup}

Unless otherwise specified, all inference is run using the \texttt{transformers} library with 16-bit floating point precision using \textsc{Flash Attention 2} \citep{dao2023flashattention2}. 
We train the DPO-based models on four NVIDIA A100-SXM4-80GB GPUs, with each model requiring approximately 2$\sim$3 days of training. 
For other experiments, including inference and evaluation using small-scale language models, we use a variable number of NVIDIA A100-SXM4-80GB GPUs depending on the model size.
For completeness, we also provide the prompt instruction for the zero-shot inference setting in Figure~\ref{fig:zero_shot_prompt}.

\paragraph{DPO Training.}
We train our DPO-based models using 4 batches and the default hyperparameter settings from the \texttt{DPOConfig} class in the \texttt{transformers} library. 
This corresponds to using an (adaptive) learning rate of $5.0\times 10^{-5}$, a $\beta$ value of $0.1$, and reverse KL divergence for $f$-divergence regularization.

\vspace{-1mm}
\paragraph{PINE.}  
We use the codebase available at \url{github.com/wzq016/PINE.git} for the PINE implementation. In our setup, the input is formatted as  
\begin{equation*}
    [\texttt{INS} \mid d_{t, \theta_1}^{(1)} \mid d_{t, \theta_1}^{(2)} \mid \ldots \mid d_{t, \theta_2}^{(1)} \mid d_{t, \theta_2}^{(2)} \mid \ldots \mid \texttt{EOS}],
\end{equation*}
where $\texttt{INS}$ is the prompt instruction, $D_{t,\theta_1} = d_{t, \theta_1}^{(1)} \mid d_{t, \theta_1}^{(2)} \mid \ldots$ represents the left-leaning source documents, $D_{t,\theta_2} = d_{t, \theta_2}^{(1)} \mid d_{t, \theta_2}^{(2)} \mid \ldots$ the right-leaning source documents, and $\texttt{EOS}$ is the end-of-sequence token. 
PINE reformats the input by designating a target segment (e.g., $D_{t,\theta_1}$ when the target perspective is the left-leaning view) to ensure that all segments are attended to uniformly, regardless of their original positions.

\vspace{-1mm}
\paragraph{Dataset.} 
We use the \polisum dataset \cite{deas2025summarization} as our primary testbed for perspective summarization. 
We remove duplicates from the dataset and obtain 1816 article pairs (left- and right-leaning). 
We split the data into 1716 article pairs for training DPO+RR and 100 article pairs for testing. 
Although most methods we investigate do not rely on training, we maintain a strict separation between train and test sets to avoid inflating DPO+RR performance.

\subsection{Ranking Methods}
\label{app:ranking_details}

To obtain accurate ranking results using automated metrics, we fit a Bradley-Terry model \cite{bradley1952rank} to the per-method scores for each bootstrap resample of the test set and derive confidence intervals for each method's ability estimate. 

Specifically, let there be $M$ methods with latent abilities $\{\theta_1,\theta_2,\ldots,\theta_M\}$. 
For each pair of methods $(i,j)$, the model posits that the probability of $i$ "winning" over $j$ in a pairwise comparison is given by a logistic function:
\vspace{-2mm}
\begin{equation}
\label{eq:bt-prob}
\Pr[\text{$i$ beats $j$}] = \frac{1}{1 + \exp\bigl(-(\theta_i - \theta_j)/\sigma\bigr)},
\end{equation}
where $\sigma > 0$ is a noise or scale parameter. We treat method $i$ as having beaten method $j$ if $i$'s aggregated raw score exceeds $j$'s, resolving exact ties randomly.

We estimate the abilities by maximizing the log-likelihood of all observed pairwise outcomes:
\begin{equation*}
\resizebox{\columnwidth}{!}{$
    \begin{aligned}
        \ell(\theta_1,\ldots,\theta_M)
        &=\sum_{(i,j)\in\mathcal{D}}
        \Bigl[
        \mathbbm{1}[i \text{ beats } j]\log \Pr[i \text{ beats } j]
        \nonumber\\
        &\quad+\;
        \mathbbm{1}[j \text{ beats } i]\log\bigl(1 - \Pr[i \text{ beats } j]\bigr)
        \Bigr],
    \end{aligned}
    $}
\end{equation*}
where $\mathcal{D}$ denotes all pairwise comparisons from the current (re)sample, and $\mathbbm{1}(\cdot)$ is an indicator function. 
We perform this fitting procedure via numerical optimization (L-BFGS). 
To account for variability, we employ bootstrap resampling over the test set: 
each resample draws the test documents (with replacement), averages each method's raw scores within that resample, and re-fits the Bradley-Terry model to generate a new set of abilities $\{\theta_m\}$. 
We repeat for $B=500$ iterations and obtain an empirical distribution of ability estimates for each method. 
We then rank methods by their mean estimated ability across all bootstrap replicates and derive $95\%$ confidence intervals from the resulting bootstrap distributions. 

\vspace{-2mm}
\section{Metric Evaluation}
\label{app:metric_evaluation}
\vspace{-2mm}

Here, we provide supplementary content for benchmarking evaluation metrics for measuring coverage and faithfulness.

\begin{table*}[t]
    \centering
    \resizebox{\linewidth}{!}{
    \begin{tabular}{l|l|cc|cc}
         \toprule
         \bf Metric & \bf Model & \multicolumn{2}{c|}{\bf Coverage} & \multicolumn{2}{c}{\bf Faithfulness} \\
                    &           & \bf Corr.~($\rho_s$) & \bf Winrate & \bf Corr.~($\rho_s$) & \bf Winrate \\
         \midrule
         \multirow{7}{*}{\llmcov} 
             & \texttt{Mistral-7B-Instruct-v0.3}          & \cellcolor{orange!70}$0.707^{***}$             & \cellcolor{orange!45}$0.739 \pm 0.047$             & \cellcolor{orange!39}\color{gray}$0.393^{***}$ & \cellcolor{orange!10}\color{gray}$0.431 \pm 0.115$ \\
             & \texttt{Mixtral-8x7B-Instruct-v0.1}        & \cellcolor{orange!70}$0.720^{***}$             & \cellcolor{orange!45}$0.771 \pm 0.050$             & \cellcolor{orange!28}\color{gray}$0.335^{***}$ & \cellcolor{orange!10}\color{gray}$0.475 \pm 0.087$ \\
             & \texttt{Llama-3.1-8B-Instruct}             & \cellcolor{orange!40}$0.606^{***}$             & \cellcolor{orange!40}$0.648 \pm 0.051$             & \cellcolor{orange!15}\color{gray}$0.188^{***}$ & \cellcolor{orange!5} \color{gray}$0.313 \pm 0.093$ \\
             & \texttt{Llama-3.3-70B-Instruct}            & \cellcolor{orange!70}$0.724^{***}$             & \cellcolor{orange!45}$0.753 \pm 0.058$             & \cellcolor{orange!28}\color{gray}$0.280^{***}$ & \cellcolor{orange!10}\color{gray}$0.415 \pm 0.100$ \\
             & \texttt{Qwen2.5-7B-Instruct}               & \cellcolor{orange!50}$0.650^{***}$             & \cellcolor{orange!45}$0.624 \pm 0.081$             & \cellcolor{orange!39}\color{gray}$0.343^{***}$ & \cellcolor{orange!10}\color{gray}$0.349 \pm 0.106$ \\
             & \texttt{Qwen2.5-14B-Instruct}              & \cellcolor{orange!70}$0.732^{***}$             & \cellcolor{orange!45}$0.749 \pm 0.049$             & \cellcolor{orange!33}\color{gray}$0.334^{***}$ & \cellcolor{orange!10}\color{gray}$0.380 \pm 0.081$ \\
             & \texttt{Qwen2.5-32B-Instruct}              & \cellcolor{orange!70}$0.721^{***}$             & \cellcolor{orange!45}$0.709 \pm 0.060$             & \cellcolor{orange!28}\color{gray}$0.302^{***}$ & \cellcolor{orange!10}\color{gray}$0.343 \pm 0.097$ \\
         \midrule
         \multirow{7}{*}{\llmfaith} 
             & \texttt{Mistral-7B-Instruct-v0.3}          & \color{gray}\cellcolor{orange!50}$0.504^{***}$ & \color{gray}\cellcolor{orange!18}$0.494 \pm 0.061$ & \cellcolor{orange!64}$0.646^{***}$             & \cellcolor{orange!29}$0.498 \pm 0.113$ \\
             & \texttt{Mistral-Large-Instruct-2411}       & \color{gray}\cellcolor{orange!50}$0.722^{***}$ & \color{gray}\cellcolor{orange!18}$0.688 \pm 0.076$ & \cellcolor{orange!64}$0.579^{***}$             & \cellcolor{orange!29}$0.479 \pm 0.108$ \\
             & \texttt{Llama-3.1-8B-Instruct}             & \color{gray}\cellcolor{orange!40}$0.577^{***}$ & \color{gray}\cellcolor{orange!14}$0.303 \pm 0.074$ & \cellcolor{orange!28}$0.439^{***}$             & \cellcolor{orange!5} $0.188 \pm 0.079$ \\
             & \texttt{Llama-3.3-70B-Instruct}            & \color{gray}\cellcolor{orange!40}$0.558^{***}$ & \color{gray}\cellcolor{orange!14}$0.283 \pm 0.080$ & \cellcolor{orange!73}$0.735^{***}$             & \cellcolor{orange!5} $0.343 \pm 0.112$ \\
             & \texttt{Qwen2.5-7B-Instruct}               & \color{gray}\cellcolor{orange!50}$0.589^{***}$ & \color{gray}\cellcolor{orange!18}$0.536 \pm 0.064$ & \cellcolor{orange!64}$0.644^{***}$             & \cellcolor{orange!29}$0.503 \pm 0.087$ \\
             & \texttt{Qwen2.5-14B-Instruct}              & \color{gray}\cellcolor{orange!70}$0.702^{***}$ & \color{gray}\cellcolor{orange!45}$0.671 \pm 0.099$ & \cellcolor{orange!60}$0.616^{***}$             & \cellcolor{orange!29}$0.519 \pm 0.086$ \\
             & \texttt{Qwen2.5-32B-Instruct}              & \color{gray}\cellcolor{orange!70}$0.712^{***}$ & \color{gray}\cellcolor{orange!45}$0.675 \pm 0.063$ & \cellcolor{orange!60}$0.670^{***}$             & \cellcolor{orange!29}$0.590 \pm 0.096$ \\
         \bottomrule
     \end{tabular}
    }
    \caption{Comparison of Spearman rank correlation ({\bf Corr.~($\rho_s$)}) and Winrate ({\bf Winr.}) across different backbone models. 
    \llmcov exhibits moderate to high correlation and winrate across all backbones, while \mistral and \qwenfourteen achieve the best performance for faithfulness. }
    \label{tab:llm_metric_evaluation}
\end{table*}

\begin{figure*}[t]
    \centering
    \begin{subfigure}[t]{0.43\linewidth}
        \centering
        \begin{lstlisting}[basicstyle=\ttfamily\scriptsize, 
                           frame=single, 
                           breaklines=true,
                           breakatwhitespace=false,
                           columns=fullflexible,
                           keepspaces=true,
                           xleftmargin=0pt,
                           xrightmargin=0pt,
                           breakindent=0ex]
You are an evaluator. Your task is to determine how well a generated summary captures all of the main arguments from a source article. This is a measure of "coverage," which does not necessarily address factual accuracy (faithfulness) but focuses on completeness of content.

The scale for coverage is:
1. No Coverage: The summary does not include any of the main arguments from the article.
2. Low Coverage: The summary includes only a few of the main arguments from the article, omitting most.
3. Medium Coverage: The summary contains around half of the article's main arguments.
4. High Coverage: The summary contains most of the main arguments from the article, missing only a few.
5. Perfect Coverage: The summary includes all major points and arguments mentioned in the article, leaving out nothing important.

Follow these steps carefully:

1. **Read the Source Article**: Examine the text provided in the article. Identify all major points, arguments, or facts it contains.
2. **Read the Summary**: Look at the text in the summary. List each argument or point the summary includes.
3. **Compare for Completeness**:
- Check if each major point from the source article is present in the summary.
- Count how many major points are covered versus how many are omitted.
4. **Determine the Score**:
- Assign a score from 1 (no coverage) to 5 (perfect coverage), based on how many main arguments are included in the summary relative to the source.
5. **Output Instructions**:
- Output only the final numeric score (1, 2, 3, 4, or 5) without any explanation or additional text.

---

# Source Article:

(article)

# Generated Summary:

(summary)

# Final Coverage Score (1~5 only):\end{lstlisting}
    \caption{Full prompt instructions for \llmcov.}
    \label{fig:llmfaith_prompt}
    \end{subfigure}
    \hfill
    \begin{subfigure}[t]{0.50\linewidth}
        \begin{lstlisting}[basicstyle=\ttfamily\scriptsize, 
                           frame=single, 
                           breaklines=true,
                           breakatwhitespace=false,
                           columns=fullflexible,
                           keepspaces=true,
                           xleftmargin=0pt,
                           xrightmargin=0pt,
                           breakindent=0ex]
You are an evaluator. Your task is to analyze how faithfully a generated summary represents the information found in the source article. Faithfulness here means the absence of factual errors---i.e., any claims in the summary must be either directly stated, heavily implied, or logically entailed by the source article.

The scale for faithfulness is:
1. Unfaithful: The summary is almost entirely incorrect or unrelated to the source.
2. Mostly Unfaithful: The summary includes only a few relevant arguments or correct details but is largely incorrect or missing.
3. Neutral: The summary has about half of the important points correct, but also includes notable inaccuracies or omissions.
4. Mostly Faithful: The summary reflects most of the article's arguments accurately, with only minor errors or omissions.
5. Perfectly Faithful: The summary includes all of the article's main arguments accurately and does not introduce any contradictory or unsupported claims.

Follow these steps carefully:

1. **Read the Source Article**: Examine the text provided in the article. Identify the main points, arguments, or facts it contains.
2. **Read the Summary**: Look at the text in the summary. Itemize or note each claim or statement made in the summary.
3. **Compare for Accuracy**:
- Check if each claim in the summary is explicitly or logically supported by the source. 
- Mark any claim that appears to be contradicting the source or not found in the source. 
- Check if the summary omits major arguments that are central to the source.
4. **Determine the Score**: 
- Assign a score from 1 (completely unfaithful) to 5 (perfectly faithful), based on how many claims match (and do not contradict) the source article and whether key points are included.
5. **Output Instructions**:
- Output only the final numeric score (1, 2, 3, 4, or 5) without any additional explanation or text.

---

# Source Article:

(article)

# Generated Summary:

(summary)

# Final Faithfulness Score (1~5 only):\end{lstlisting}
    \caption{Full prompt instructions for \llmfaith.}
    \label{fig:llmcov_prompt}
    \end{subfigure}
    \caption{Complete prompt instructions for both attributes in prompting-based scoring. 
    The model is provided with descriptions of each score value and a step-by-step procedure for evaluating the summary based on the article.}
    \label{fig:llmscoring_prompts}
    \vspace{-4mm}
\end{figure*}

\begin{figure*}[t]
    \centering
    \begin{subfigure}[t]{0.36\linewidth}
        \centering
        \begin{lstlisting}[basicstyle=\ttfamily\scriptsize, 
                           frame=single, 
                           breaklines=true,
                           breakatwhitespace=false,
                           columns=fullflexible,
                           keepspaces=true,
                           xleftmargin=0pt,
                           xrightmargin=0pt,
                           breakindent=0ex]
[TASK]
You are given an article that makes an argument related to the provided topic. An excerpt from the document highlights the main key argument that the author of the article is trying to assert. Please write a concise, short, one-sentence paraphrase (as short as possible) that reflects the argument implied or present in the provided excerpt. **Your paraphrase should begin with "The article argues"**.

---

Topic: (topic)

Article: (article)

Excerpt: (excerpt)

---

One-Line Argument Summary starting with "The article argues":\end{lstlisting}
    \vspace{-3mm}
    \caption{Full prompt instructions for paraphrasing highlighted excerpts to key points.}
    \label{fig:paraphrase_prompt}
    \vspace{-2mm}    
    \end{subfigure}
    \hfill
    \begin{subfigure}[t]{0.58\linewidth}
        \begin{lstlisting}[basicstyle=\ttfamily\scriptsize, 
                           frame=single, 
                           breaklines=true,
                           breakatwhitespace=false,
                           columns=fullflexible,
                           keepspaces=true,
                           xleftmargin=0pt,
                           xrightmargin=0pt,
                           breakindent=0ex]
[TASK]
You are given one main argument from a political news article (either left-leaning or right-leaning). **Rewrite the argument so that the argument is completely reversed or semantically opposite.** If the original argument supports or praises a policy/idea/group, the reversed version should criticize or oppose it, and vice versa. Only return the reversed argument itself, with no extra commentary or analysis.

[EXAMPLES]
1.
ORIGINAL: The article argues that stricter immigration laws help protect domestic jobs and strengthen national identity.
REVERSED: The article argues that relaxed immigration laws create more job opportunities and enhance cultural diversity.

2.
ORIGINAL: The article insists that climate change is primarily caused by human activity and demands immediate government intervention.
REVERSED: The article insists that human activity has minimal impact on climate change and calls for minimal government involvement.

[INFERENCE]
ORIGINAL: (original key point)
REVERSED:\end{lstlisting}
    \vspace{-3mm}
    \caption{Prompt for generating adversarial key points $\overline{K}_{t,\theta}$ from the curated key points ${K}_{t,\theta}$.}
    \label{fig:adversarial_prompt}
    \vspace{-2mm}    
    \end{subfigure}
    \caption{Prompts used for portions of the procedure for curating the benchmarking test set for metrics.}
\end{figure*}

\begin{figure*}[t]
    \centering
    \begin{subfigure}[t]{0.49\textwidth}
        \centering
        \includegraphics[width=\linewidth]{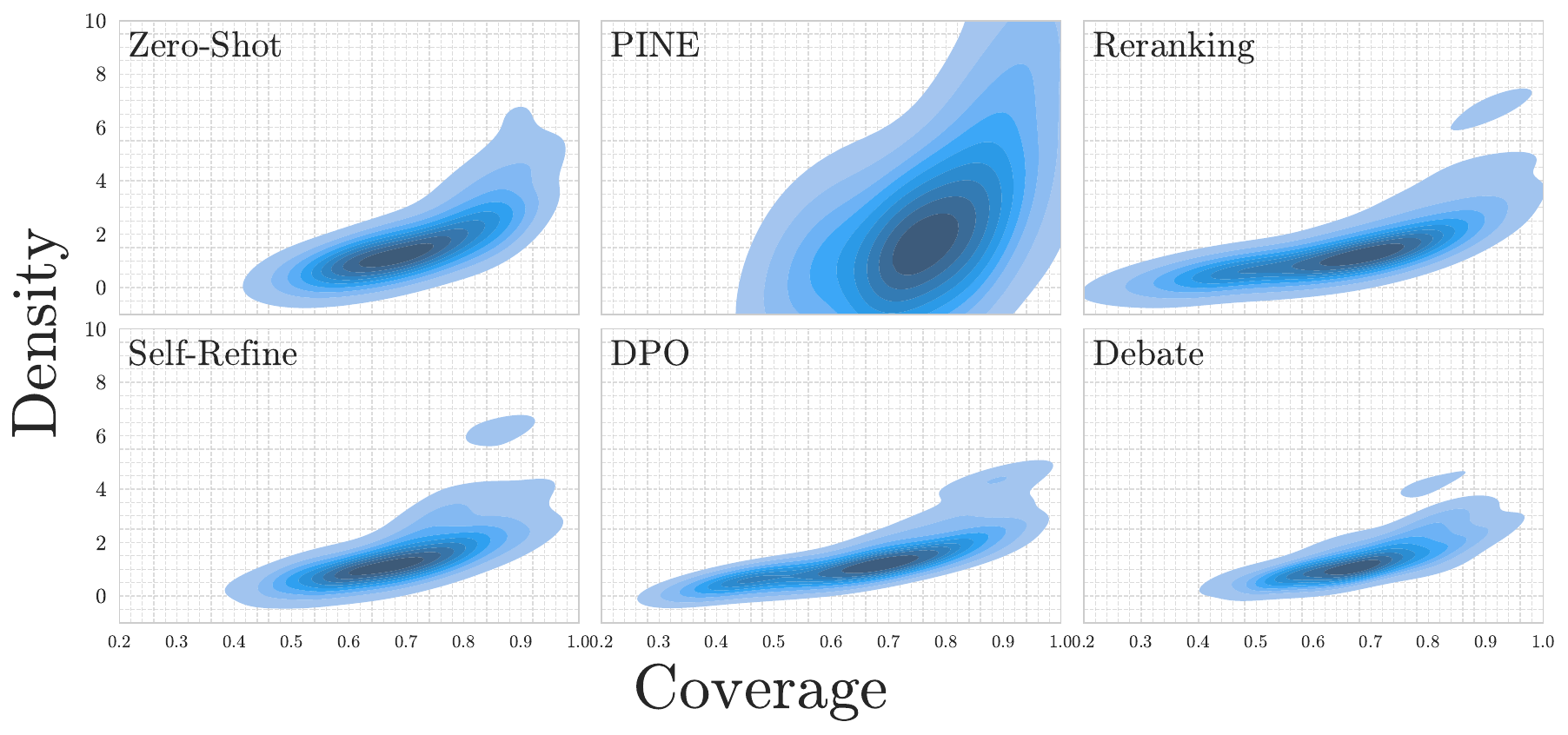}
        \caption{Coverage-Density plots for the source document.}
        \label{fig:density_source}
        \vspace{-2mm}                
    \end{subfigure}
    \begin{subfigure}[t]{0.49\textwidth}
        \centering
        \includegraphics[width=\linewidth]{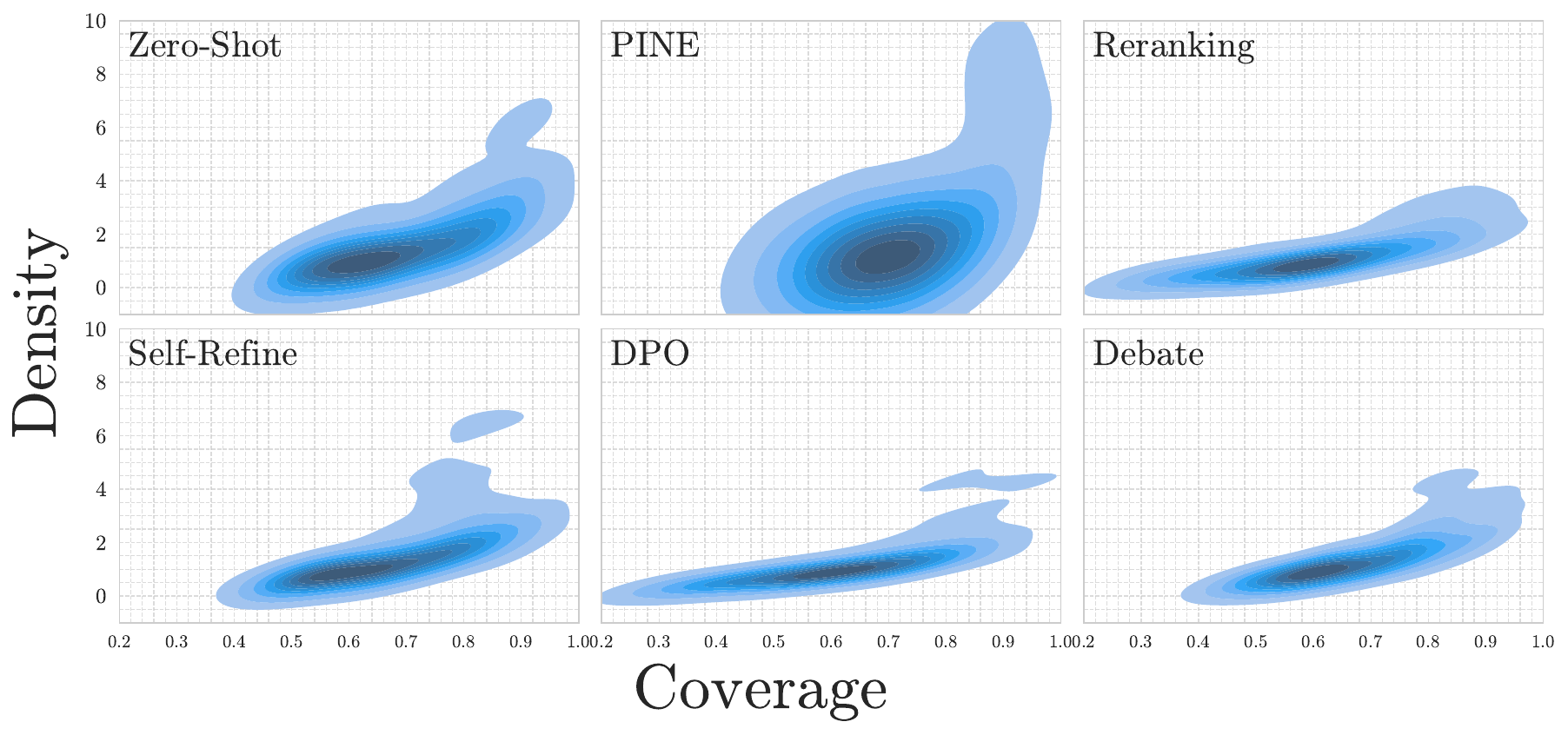}
        \caption{Coverage-Density plots for the opposing document.}
        \label{fig:density_opposite}
        \vspace{-2mm}        
    \end{subfigure}
    \caption{Coverage-density plots for source and opposing documents.
    PINE exhibits higher variance in coverage, while other methods follow a similar structure.}
    \label{fig:density_plots}
    \vspace{-3mm}
\end{figure*}

\begin{table}[t]
    \centering
    \resizebox{\columnwidth}{!}{
    \begin{tabular}{l|lcc}
    \toprule
    \bf Method        & \bf Summary Length & \bf EF Coverage    & \bf Comp.\ Ratio \\
    \midrule\midrule 
    Zero-Shot         & \cellcolor{orange!52} $40.77 \pm 6.212 $ & \cellcolor{orange!53} $0.719 \pm 0.113$ & \cellcolor{orange!29} $14.958 \pm 4.165$ \\
    Self-Refine       & \cellcolor{orange!80} $43.94 \pm 10.73 $ & \cellcolor{orange!40} $0.692 \pm 0.107$ & \cellcolor{orange!30} $15.097 \pm 5.774$ \\
    Debate            & \cellcolor{orange!58} $41.50 \pm 11.609$ & \cellcolor{orange!40} $0.692 \pm 0.103$ & \cellcolor{orange!42} $16.192 \pm 4.903$ \\
    PINE              & \cellcolor{orange!29} $38.17 \pm 7.401 $ & \cellcolor{orange!80} $0.776 \pm 0.125$ & \cellcolor{orange!80} $19.589 \pm 20.23$ \\
    Reranking         & \cellcolor{orange!20} $37.13 \pm 13.856$ & \cellcolor{orange!20} $0.651 \pm 0.157$ & \cellcolor{orange!20} $14.166 \pm 4.181$ \\
    DPO+RR            & \cellcolor{orange!71} $42.94 \pm 8.245 $ & \cellcolor{orange!25} $0.661 \pm 0.149$ & \cellcolor{orange!22} $14.391 \pm 4.421$ \\
    \bottomrule
    \end{tabular}
    }
    \caption{Supplementary statistics for each method, measured by summary length, extractive fragment coverage, and compression ratio.}
    \label{tab:summary_abstractiveness_all}
    \vspace{-6mm}
\end{table}

\vspace{-2mm}
\subsection{Metric Configurations}
\label{app:metric_configurations}

For \rouge, we use the \texttt{rouge-score} Python library. 
For \bert and \bleurt, we use the \texttt{deberta-large-xnli} and \texttt{BLEURT-20-D6} checkpoints respectively, due to their higher correlations with human judgments. 
For \alignscore, we employ the \texttt{AlignScore-large} checkpoint from \citet{zha2023alignscore}. 
For \summac, we use the \texttt{tals/albert-xlarge-vitaminc-mnli} model, which is the default setting for the \summac evaluation metric.

\vspace{-2mm}
\subsection{Prompt Instructions}
\label{app:metric_annotation_prompts}

\paragraph{Prompt-based Scoring: \llmcov and \llmfaith.}
We provide the full prompt instructions for both \llmcov and \llmfaith in Figures~\ref{fig:llmcov_prompt} and \ref{fig:llmfaith_prompt}, respectively. 
While we experiment with prompt variations such as using binary and ternary scoring and removing step-by-step procedures, these modifications result in lower performance. 
We omit these alternate prompts for brevity.

\vspace{-2mm}
\subsection{Backbone Scoring Evaluation}
\label{app:llm_metric_benchmarking}
\vspace{-1mm}

Table~\ref{tab:llm_metric_evaluation} presents additional results for various LLM backbones. 
Notably, prompt-based scoring generally performs better on coverage than on faithfulness. 
In particular, \mistral and \qwenfourteen exhibit the best performance across both metrics. 
Based on these results, we use \mistral as the evaluator and \qwenfourteen as the proxy metric.
For coverage, larger model sizes weakly correlate with higher performance, though the gains are marginal. 
To keep inference time reasonable, we therefore use smaller-scale models that still exhibit good performance. 
For faithfulness, the \texttt{Llama} models consistently underperform compared to other backbones on both correlation and ranking. 
However, as all backbones perform close to the random baseline on winrate, we avoid using prompt-based scoring for faithfulness.

\vspace{-2mm}
\subsection{Paraphrasing Excerpts to Key Points}
\label{app:paraphrasing}
\vspace{-1mm}

As mentioned in \S\ref{sec:assessing_metric_quality}, we use an LLM to paraphrase highlighted excerpts into key points. 
We also employ an LLM to generate adversarial key points $\overline{K}_{t,\theta}$ from the curated key points $K_{t,\theta}$, using the prompts provided in Figures~\ref{fig:paraphrase_prompt} and \ref{fig:adversarial_prompt}. 
We use \texttt{Qwen2.5-32B-Instruct} for both paraphrasing and key point generation.

\vspace{-3mm}
\section{Benchmarking Methods}
\vspace{-2mm}

Here, we provide supplementary details on our evaluation procedure along with additional analysis on the generated summaries by each method.

\vspace{-2mm}
\subsection{Inter-Annotator Agreement}
\label{app:iaa_details}

We first provide additional information on the matching function $M(\cdot, \cdot)$ in \S\ref{sec:results_human_evaluation}. 
For each element $s_i^A \in S_A$, the function finds the first unmatched element $s_j^B \in S_B$ that meets a matching condition.
The first criterion is exact containment: if $s_i^A$ is a substring of $s_j^B$ or vice versa, they are considered a match. 
If no exact containment is found, we compute the longest common subsequence (LCS) between $s_i^A$ and $s_j^B$.
If the LCS length divided by the length of the shorter string exceeds a predefined threshold $\tau$, we consider them a match. 
Each element in $S_A$ is matched to at most one element in $S_B$, and vice versa, and is removed from further matching once paired. 
By default, we set $\tau = 0.5$.

\vspace{-2mm}
\paragraph{Random Baseline for IAA.}
We simulate random highlight selection as follows. 
First, we compute the mean and variance of the number of highlights and their lengths separately for documents and summaries. 
Using these statistics, we sample the number of highlights and the length of each highlight from a normal distribution $\mathcal{N}(\cdot, \cdot)$ for each document or summary. 
We repeat this process independently twice and compute the overlap between the two instances as described in \S\ref{sec:results_human_evaluation}. 
This procedure simulates a non-trivial, semi-realistic random highlighting of excerpts in documents and summaries.

\subsection{Supplementary Analysis}
\label{app:methods_supplementary_analysis}

In addition to coverage-density plots, we report additional results for summary lengths, \emph{extractive fragment coverage} (quantifying the extent of copying from the source), and \emph{compression ratios} \cite{grusky2018newsroom} (assessing summary length relative to the source document). 

Table~\ref{tab:summary_abstractiveness_all} shows our results.
Consistent with findings in \S\ref{sec:summary_characteristics}, PINE exhibits lower abstractiveness compared to other methods.
Moreover, both compression ratios and summary lengths indicate that PINE tends to generate shorter summaries relative to other methods.

\paragraph{Extractive Fragment Plots.}

We also include Coverage-Density plots for both the source and opposing perspective documents in Figure~\ref{fig:density_plots}.
Overall, we observe similar coverage-density structures for all by PINE, which exhibits a wider spectrum of coverage and density. 
This indicates that the abstractiveness of PINE exhibits high variance, whereas for other methods the abstractiveness is relatively stable.
Furthermore, we also see that the coverage-density plots for the opposing side is slightly lower than for the target source articles.

\vspace{-3mm}
\subsection{Additional Example Summaries}
\label{app:example_summaries}
In Table~\ref{tab:example_summaries_app}, we provide additional sampled examples for the summaries generated by each method. 

\begin{table*}[t]
    \centering
    \scriptsize
    \begin{tabularx}{\linewidth}{l|X}
        \toprule
        \bf Topic & \bf Student Loans\\
        \midrule
        \midrule
        Zero-Shot & The Right argues that Biden's student loan relief plan disproportionately benefits those making less than \$75,000 per year and that it is a Band-Aid solution that fails to address the root cause of the issue, which is the increasing cost of higher education due to colleges and universities jacking up their prices. \\
        \midrule        
        Self-Refine & The Right argues that President Biden's student loan relief plan benefits low-income individuals, excludes the truly rich, and is not a giveaway to the elite, while the root cause of the student debt crisis lies in government subsidies to colleges, which have led to inflated tuition prices. \\
        \midrule        
        Debate & The Right argues that President Biden's student loan relief plan is a reasonable measure to help lower-income individuals, who make up 90\% of those benefiting, due to the government's subsidization of colleges and universities leading to inflated tuition costs. \\
        \midrule        
        PINE & The Right argues that the student debt relief plan is a Band-Aid that will collapse and that colleges should be expected to finance the education they provide.\\
        \midrule        
        Reranking & The Right suggests that Biden's student debt relief plan is a Band-Aid solution that primarily benefits those making less than \$75,000 per year and does not address the root cause of rising college costs, which can be mitigated by requiring institutions to cut costs and improve their financial efficiency. \\
        \midrule        
        DPO+RR & The Right views Biden's student-loan debt forgiveness plan as a Band-Aid solution that unfairly excludes the truly rich and argues that colleges and universities should be held accountable for their role in inflating prices and should provide debt-free education to students. \\
        \bottomrule
    \end{tabularx}
    
    \vspace{1em}
    
    \begin{tabularx}{\linewidth}{l|X}
        \toprule
        \bf Topic & \bf Mitt Romney\\
        \midrule
        \midrule
        Zero-Shot & The Left sees the failure of Republicans to convict Trump as a decision made "functionally at gunpoint" due to the perceived threat of violence from the extremist right, rather than through conviction or political calculation. \\
        \midrule        
        Self-Refine & The Left views the failure of Republicans to convict Trump as a result of the perceived violent threat from the extremist right, which swayed their decision-making due to fear for their families' safety, representing one of the most consequential decisions in American history. \\
        \midrule        
        Debate & The Left views Romney's decision to stay in the Republican Party as a failure to challenge the extremist right and uphold democratic values, having been swayed by perceived threats of violence rather than conviction or principle. \\
        \midrule        
        PINE & The Left believes that Mitt Romney's decision to leave the Republican Party is a testament to his commitment to principle and his willingness to stand up to the party's extremist elements.\\
        \midrule        
        Reranking & The Left portrays Romney as revealing a stark contrast between his authentic disdain for the extreme ideologies within his own party in private conversations, versus his careful and cautious language used in public statements, highlighting the profound impact of fear generated by far-right extremism on congressional decision-making processes, specifically in relation to Trump's impeachment trial and acquittal. \\
        \midrule        
        DPO+RR & The Left views Senator Romney's decision not to convict Donald Trump due to fears of potential retribution from extremists as indicative of the broader breakdown within the Republican party, illustrating its susceptibility to authoritarianism and inability to stand up against violence-based intimidation despite holding supposedly "right" convictions. \\
        \bottomrule
    \end{tabularx}
    
    \vspace{1em}
    
    \begin{tabularx}{\linewidth}{l|X}
        \toprule
        \bf Topic & \bf Democrats Sign Letter Opposing Pelosi\\
        \midrule
        \midrule
        Zero-Shot & The Left emphasizes that Nancy Pelosi's experience, legislative skill, and ability to pass significant bills, such as the Affordable Care Act, make her the ideal choice to lead the House Democrats. \\
        \midrule        
        Self-Refine & The Left believes Nancy Pelosi's experience and effectiveness make her the best choice for Speaker, despite potential risks to LGBTQ rights and ordinary women's concerns.\\
        \midrule        
        Debate & The Left emphasizes the importance of retaining Nancy Pelosi as Speaker of the House due to her extensive experience, legislative skill, and ability to pass key progressive policies, with her being a crucial figure in landmark legislative successes and there not being a clear alternative. \\
        \midrule        
        PINE & The Left argues that Nancy Pelosi is a highly effective legislator who has consistently been in the top third of most liberal Democrats in the House, and that ousting her would be a blow to ordinary women and LGBTQ people whose concerns will lose priority status if the anti-Pelosi wing prevails.\\
        \midrule        
        Reranking & The Left believes that keeping Nancy Pelosi as Speaker due to her vast experience and proven ability to effectively navigate Congress towards Democratic achievements and maintain party unity is vital amidst any challenges faced or proposed replacements who do not offer comparable qualifications or stronger alliances, especially regarding marginalized groups interests. \\
        \midrule        
        DPO+RR & The Left emphasizes that retaining Nancy Pelosi as Speaker of The House is crucial due to her extensive experience, successful legislation passing track record, ability to effectively oversee the Democratic Party, and importance in maintaining progress on key issues like LGBTQ rights and access to affordable healthcare. \\
        \bottomrule
    \end{tabularx}

    \vspace{1em}
    
    \begin{tabularx}{\linewidth}{l|X}
        \toprule
        \bf Topic & \bf China and Coronavirus\\
        \midrule
        \midrule
        Zero-Shot & The Right suggests that the United States should not completely decouple from China on public-health matters, as past collaboration has been effective, but rather demands institutional changes in China's domestic food safety policies and calls for greater economic independence from China. \\
        \midrule        
        Self-Refine & The Right believes that China's disastrous cover-up and propaganda efforts have been matched by the US's own bumbling response to the crisis.\\
        \midrule        
        Debate & The Right views China's actions as deceitful and propaganda-driven, but also acknowledges that both the US and China have mishandled the pandemic, with the US's decoupling from China on public health matters and the US's bumbling response to the pandemic exacerbating the crisis. \\
        \midrule        
        PINE & The Right argues that the United States should distance itself from China and prioritize economic independence as a permanent national security goal.\\
        \midrule        
        Reranking & The Right perspective emphasizes that the US and the world should be cautious of China's intentions and actions, particularly its propaganda efforts, and that the US should prioritize economic independence and not rely on China for aid or information, while also acknowledging that the US has made mistakes in its response to the pandemic. \\
        \midrule        
        DPO+RR & The Right emphasizes that while China failed catastrophically during the COVID-9 crisis through its aggressive disinformation campaign and inadequate initial response, the sudden shift towards dependency on China-led relief efforts actually highlights broader risks inherent in the interconnection between nations, necessitating reconsideration of international relationships and prioritization of economic sovereignty. \\
        \bottomrule
    \end{tabularx}
    
    \vspace{-3mm}
    \caption{Additional example summaries generated by each method.}
    \label{tab:example_summaries_app}
\end{table*}
\section{Annotation Information}

\vspace{-2mm}
\subsection{Annotation Details}
\label{app:annotation_details}
\vspace{-1mm}

For both annotation procedures, annotators consented to having their annotated excerpts used for research purposes (cf. Figures~\ref{fig:metric_annotation_intro} and \ref{fig:summary_annotation_intro}). 
All human evaluations in this work were conducted under an approved IRB protocol.

\paragraph{Test Set for Metric Evaluation.}
We recruited 5 graduate annotators, each assigned 10 documents for excerpt highlighting. 
Each annotator received \$15 as compensation.

\vspace{-2mm}
\paragraph{Summary Evaluation.}
Annotators were recruited from undergraduate Political Science students with self-reported knowledge of conservative and liberal beliefs to ensure the required expertise to judge summary perspectives. 
Four annotators participated—three annotated 20 documents each and one annotated 15. 
To measure inter-annotator agreement, overlapping annotations were collected for 10 documents, with each document annotated by two annotators. 
This process yielded a total of 75 document-summary annotations per method. 
Annotators were compensated at \$22.50 per hour and spent approximately $15 \pm 2.5$ minutes per page.

\subsection{Annotation Interfaces}
\label{app:annotation_interfaces}

\begin{figure*}[t]
    \centering
    \includegraphics[width=0.43\linewidth]{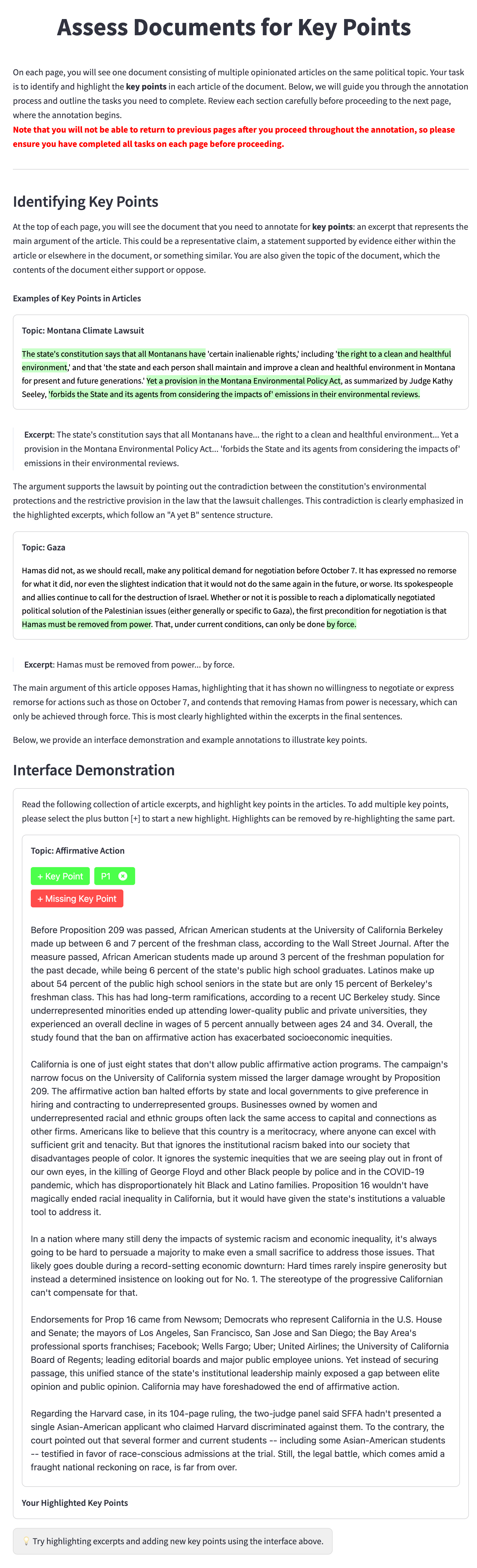}
    \includegraphics[width=0.43\linewidth]{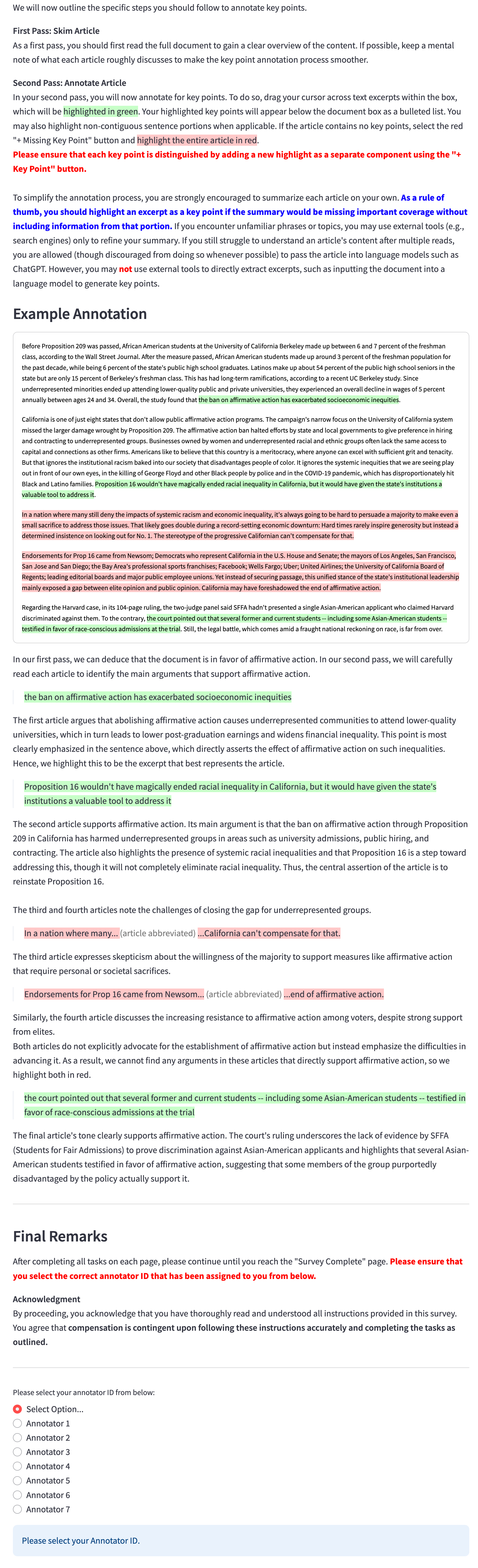}
    \caption{Introduction page for Annotation interface for annotating for article excerpts to evaluate metrics.
    Annotators are provided with definitions and an example annotated document.}
    \label{fig:metric_annotation_intro}
\end{figure*}

\begin{figure*}[t]
    \centering
    \includegraphics[width=0.8\linewidth]{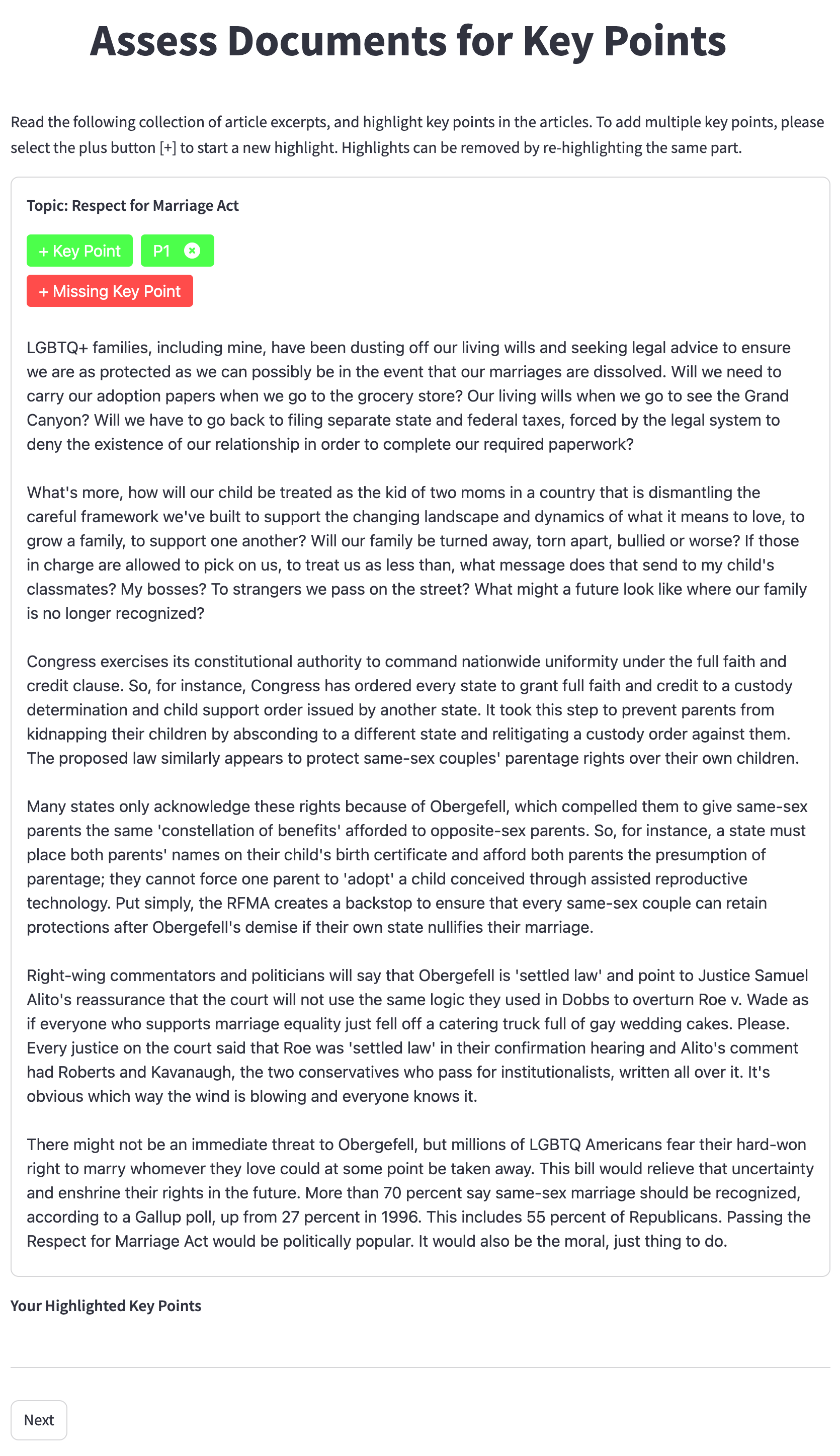}
    \caption{Example of annotation page for Annotation interface for annotating for article excerpts to evaluate metrics.
    Annotators are provided with an interface for highlighting sentences in the article.}
    \label{fig:metric_annotation_task}
\end{figure*}

\begin{figure*}[t]
    \centering
    \includegraphics[width=0.32\linewidth]{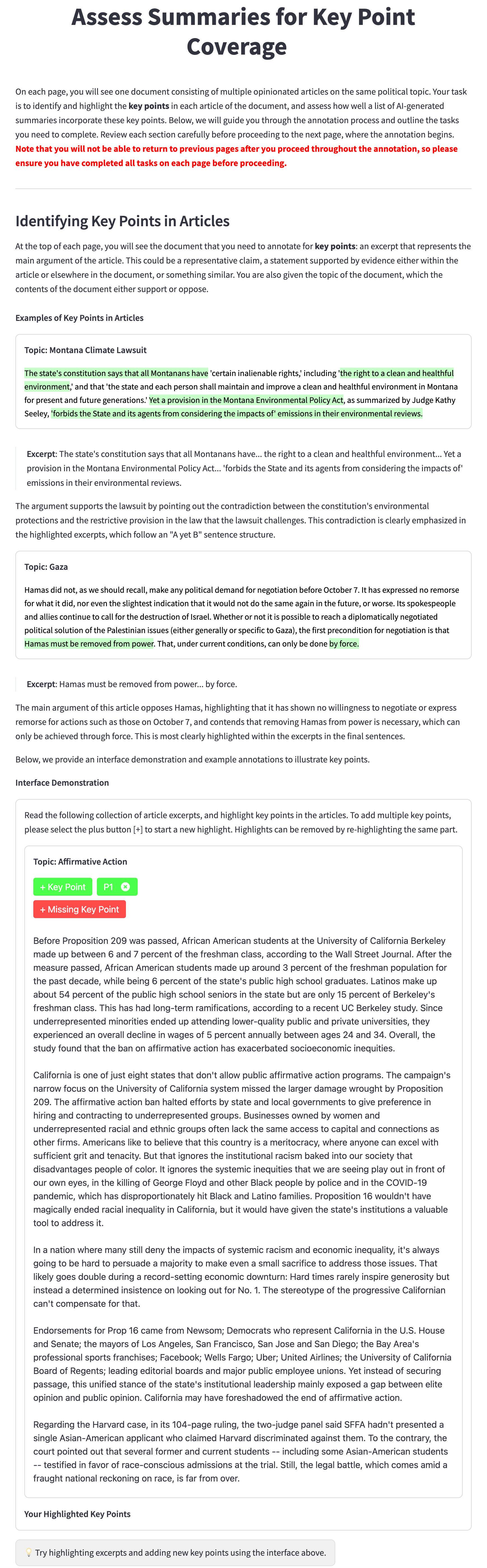}
    \includegraphics[width=0.32\linewidth]{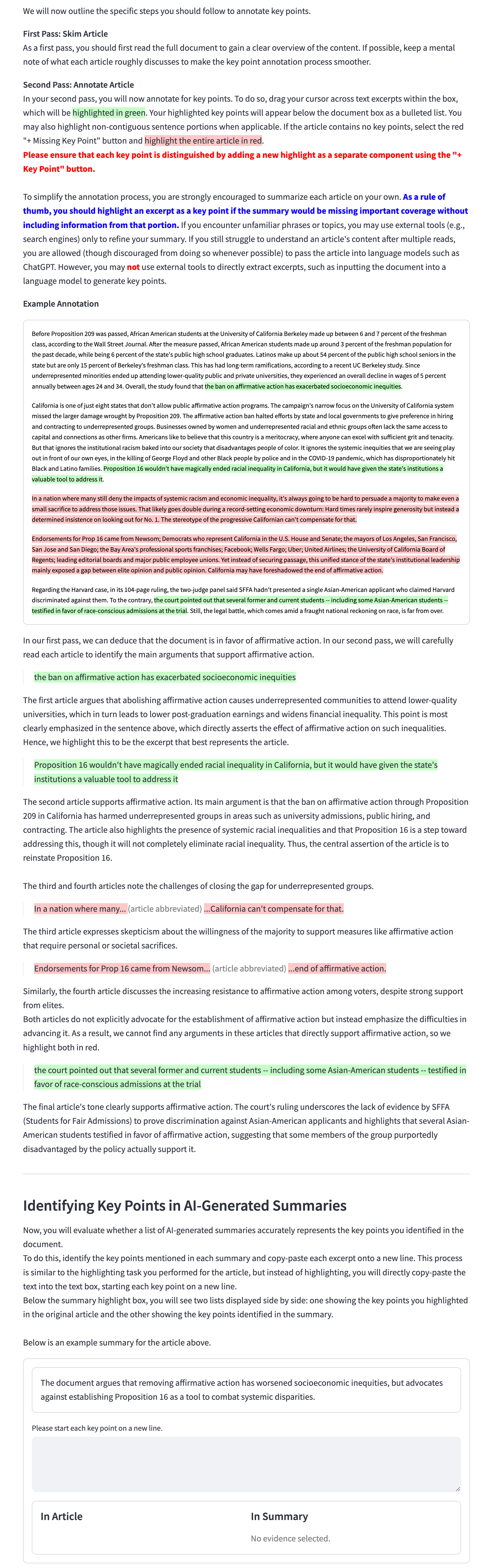}
    \includegraphics[width=0.32\linewidth]{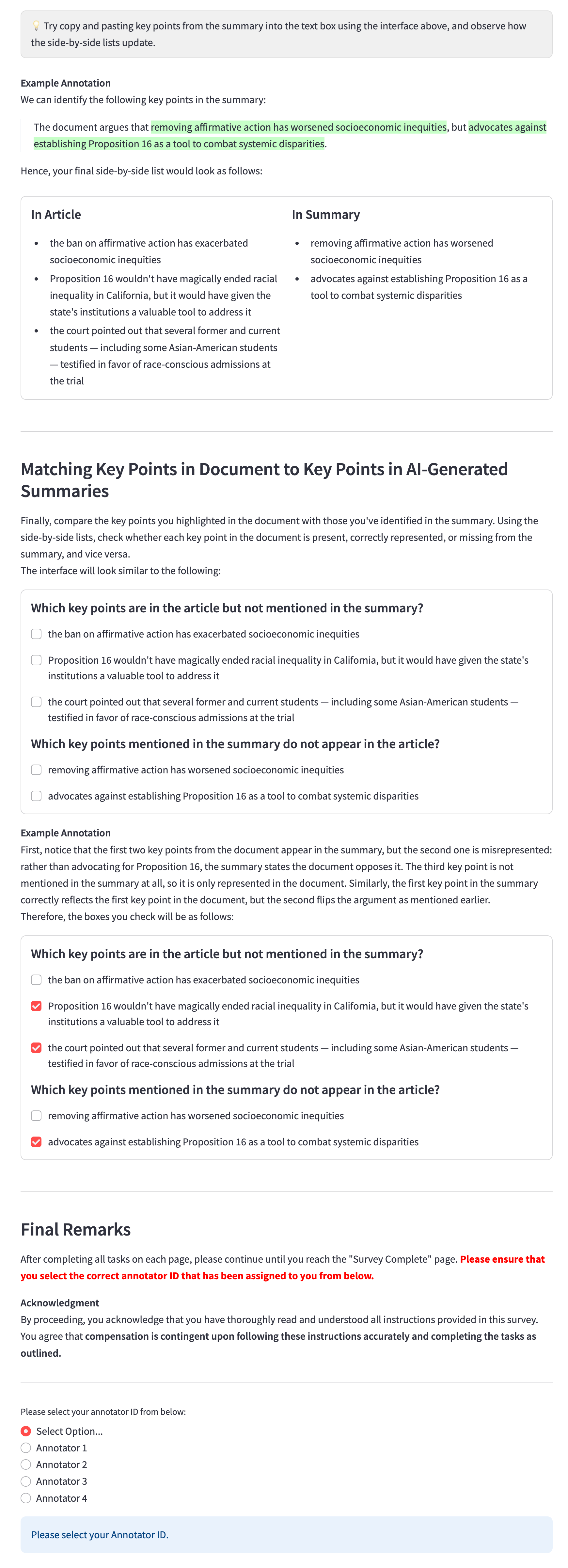}
    \caption{Introduction page of annotation interface for annotating for document and summary excerpts for evaluating method-generated summaries.}
    \label{fig:summary_annotation_intro}
\end{figure*}

\begin{figure*}[t]
    \centering
    \includegraphics[width=0.48\linewidth]{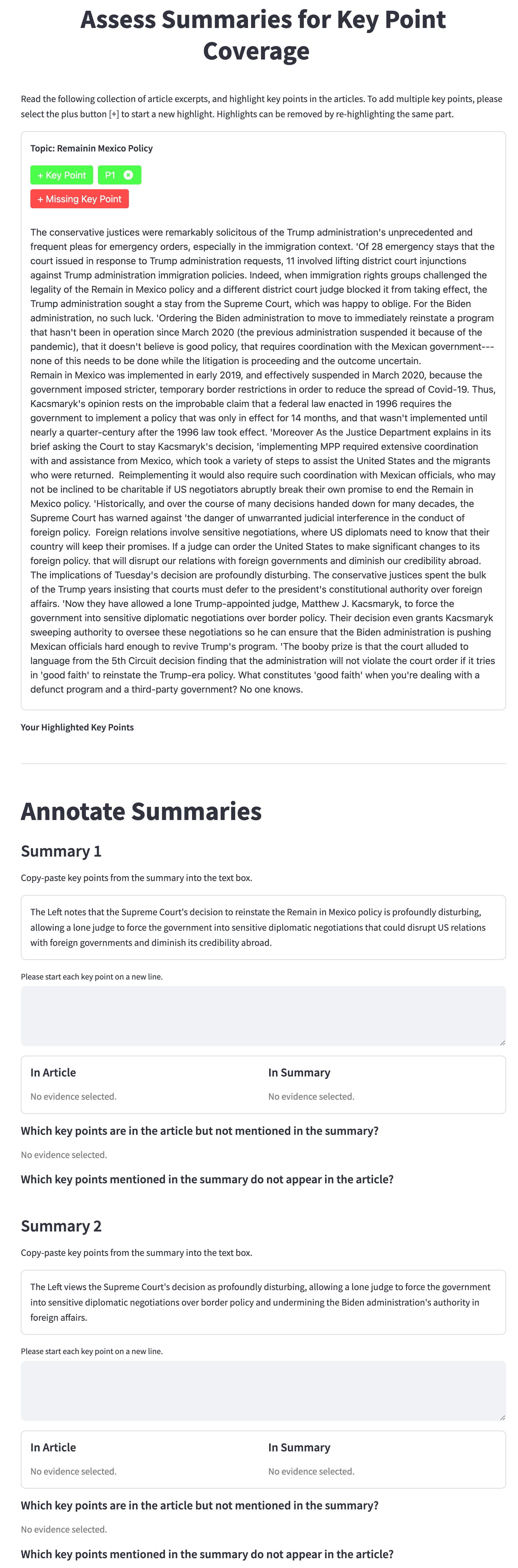}
    \includegraphics[width=0.48\linewidth]{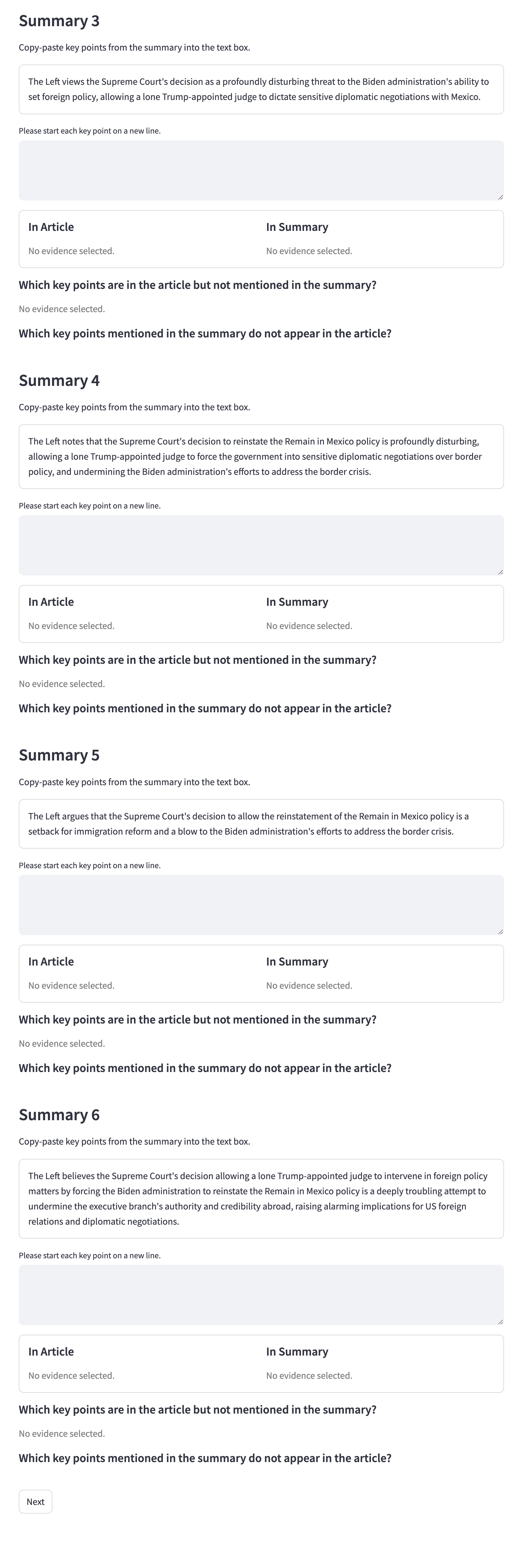}
    \caption{Example of annotation page for Annotation interface for document and summary excerpts for evaluating method-generated summaries.}
    \label{fig:summary_annotation_task}
\end{figure*}

We provide the annotation interfaces for the human studies described in \S\ref{sec:assessing_metric_quality} and \S\ref{sec:human_evaluation} in Figures~\ref{fig:metric_annotation_intro} and \ref{fig:metric_annotation_task} (for metric evaluation) and Figures~\ref{fig:summary_annotation_intro} and \ref{fig:summary_annotation_task} (for summary evaluation). 
Both interfaces were built using the \texttt{streamlit} Python library and hosted on the Streamlit Community Cloud platform.
Annotator results were stored using Amazon Web Services (AWS) Simple Storage Service (S3).

\end{document}